%% file: main.tex
\PassOptionsToPackage{table}{xcolor}
\documentclass[acmtog]{acmart}
\acmSubmissionID{1844}

\usepackage{booktabs} %
\usepackage{multirow}
\usepackage{xcolor}
\definecolor{egoplaytheme}{HTML}{E66A01}

\usepackage[ruled]{algorithm2e} %

\SetAlFnt{\small}
\SetAlCapFnt{\small}
\SetAlCapNameFnt{\small}
\SetAlCapHSkip{0pt}

\acmJournal{TOG}

\setcopyright{none}
\renewcommand\footnotetextcopyrightpermission[1]{}
\AtEndPreamble{
    \usepackage[capitalize]{cleveref}
    \crefname{section}{Sec.}{Secs.}
    \Crefname{section}{Section}{Sections}
    \Crefname{table}{Table}{Tables}
    \crefname{table}{Tab.}{Tabs.}
}

\input{command_arxiv}
\arxivbuildtrue

\makeatletter
\renewcommand\@fnsymbol[1]{\ensuremath{\dagger}}
\makeatother

\renewcommand{\datasynthref}{\cref{sec:data}, \cref{fig:data_pipe}}
\renewcommand{\additionalbaselinesref}{Table~\ref{tab:additional_baselines}}
\renewcommand{\runwayresultsref}{Table~\ref{tab:runway_results}}
\renewcommand{\longhorizonref}{Table~\ref{tab:long_horizon}}

\renewcommand{\egoeditcite}{~\cite{egoedit}}

\begin{document}

\title{EgoPlay: Event-Triggered Video Editing for Egocentric Streams}

\input{authors}
\authorsaddresses{}

\input{sections/0_abstract}



\begin{teaserfigure}
  \vspace{-0.75em}
  \makebox[\textwidth][l]{\footnotesize\textsuperscript{\(\dagger\)} Corresponding author.}
  \centerline{\large \textbf{Project page:}\quad\textcolor{egoplaytheme}{\href{https://egoplay2026.github.io/egoplay}{\underline{\texttt{https://egoplay2026.github.io/egoplay}}}}}
\end{teaserfigure}

\maketitle

\input{sections/1_intro}
\input{sections/2_related}

\input{sections/3_method}

\input{sections/4_experiment}

\input{sections/5_conclusion}
\input{sections/acknowledgments}

\input{figures/comp_figure2}
\input{figures/self_forcing}

\clearpage
\appendix
\input{sections/supp}

\bibliographystyle{ACM-Reference-Format}
\bibliography{reference}

\end{document}

%% file: command_arxiv.tex
\input{command_common}

\newcommand{\benchmarksourcecount}{150}
\newcommand{\benchmarkreportedsamples}{600}
\newcommand{\egoplayegoeditgains}{17.7\%, 16.9\%, and 16.4\%}
\newcommand{\egoplayvlmgains}{15.7\%, 14.5\%, and 13.5\%}

\newcommand{\baselineseparationtext}{We keep these differently sized evaluations separate from the 150-source main table.}

%% file: command_common.tex
\definecolor{egoplayblue}{HTML}{5F9ED1}
\definecolor{egoplayblueDark}{HTML}{1F5F99}
\definecolor{egoplayorange}{HTML}{F6A25B}
\definecolor{egoplayorangeDark}{HTML}{A94A12}
\definecolor{egoplaypanel}{HTML}{F4F6F8}
\definecolor{egoplayline}{HTML}{AEB9C7}

\makeatletter
\@ifundefined{ifsubmittedbenchmark}{\newif\ifsubmittedbenchmark\submittedbenchmarkfalse}{}
\@ifundefined{ifcamerareview}{\newif\ifcamerareview\camerareviewfalse}{}
\@ifundefined{ifarxivbuild}{\newif\ifarxivbuild\arxivbuildfalse}{}
\makeatother

\ifcamerareview
  \newcommand{\revision}[1]{{\color{blue}#1}}
\else
  \newcommand{\revision}[1]{#1}
\fi

\newcommand{\datasynthref}{the data-synthesis section and data-generation pipeline figure in the main paper}
\newcommand{\additionalbaselinesref}{the supplemental baseline table}
\newcommand{\runwayresultsref}{the commercial-system table in the supplement}
\newcommand{\longhorizonref}{the long-horizon table in the supplement}

\newcommand{\egoeditcite}{}

%% file: authors.tex
\author{Jinjie Mai}
\affiliation{%
  \institution{Snap Inc.}
  \country{USA}}
\affiliation{%
  \institution{King Abdullah University of Science and Technology (KAUST)}
  \country{Saudi Arabia}}

\author{Gordon Guocheng Qian}
\ifarxivbuild
\authornotemark[1]
\else
\authornote{Corresponding author.}
\fi
\affiliation{%
  \institution{Snap Inc.}
  \country{USA}}

\author{Willi Menapace}
\affiliation{\institution{Snap Inc.}\country{USA}}
\author{Arpit Sahni}
\affiliation{\institution{Snap Inc.}\country{USA}}
\author{Chaoyang Wang}
\affiliation{\institution{Snap Inc.}\country{USA}}
\ifarxivbuild
\author{Ashkan Mirzaei}
\affiliation{\institution{Snap Inc.}\country{USA}}
\author{Runjia Li}
\affiliation{\institution{Snap Inc.}\country{USA}}
\fi
\author{Sergey Tulyakov}
\affiliation{\institution{Snap Inc.}\country{USA}}
\author{Bernard Ghanem}
\affiliation{\institution{King Abdullah University of Science and Technology (KAUST)}\country{Saudi Arabia}}
\author{Peter Wonka}
\affiliation{\institution{Snap Inc.}\country{USA}}
\affiliation{\institution{King Abdullah University of Science and Technology (KAUST)}\country{Saudi Arabia}}
\author{Rameen Abdal}
\affiliation{\institution{Snap Inc.}\country{USA}}

\renewcommand{\shortauthors}{Mai et al.}

%% file: sections/0_abstract.tex
\begin{abstract}
\revision{We introduce \textbf{EgoPlay}, an event-triggered video-to-video editor for egocentric streams, obtained by fine-tuning a pretrained V2V diffusion transformer on event-conditioned data built primarily from Ego4D.}
Given a monocular video and an event-triggered prompt of the form ``when X happens, do Y,'' EgoPlay infers whether and when event X occurs,
preserves pre-event frames, and applies edit Y only to the post-event continuation.
Rather than cascading a separate event detector with an editor, EgoPlay learns event recognition, temporal restraint, and pixel-level editing jointly in a single end-to-end model, while also handling negative and multi-event prompts.
To support this, we construct a large-scale dataset of 106K event-triggered clip--prompt pairs spanning positive triggers, fabricated-trigger negatives, and multi-event prompts.
We then train a bidirectional video diffusion editor with event-triggered supervision and derive a causal variant for chunk-by-chunk streamable inference.
We further introduce an event-aware evaluation protocol that separately measures post-trigger editing quality, pre-trigger preservation, and false-trigger robustness.
On the Ego4D benchmark, EgoPlay substantially outperforms EgoEdit, the state-of-the-art instruction-based egocentric video editing baseline, with relative gains of \egoplayegoeditgains{} in editing quality, visual quality, and background consistency.
It also surpasses a VLM-guided detector--editor baseline by \egoplayvlmgains{} on the same metrics, while using less than half the GPU memory.
\end{abstract}

%% file: sections/1_intro.tex
\section{Introduction}
\input{figures/teaser}

Augmented reality (AR) is moving toward always-on, egocentric systems that seamlessly augment the physical world in response to wearer's behavior~\cite{grubert2017pervasive}. 
These applications require visual effects that are not merely pre-scripted, but \emph{causally driven} by what users actually do. 
A user wearing a single RGB camera should be able to specify rules such as ``\emph{when I pick up the cup, make it a goblet}'' or 
``\emph{when I upturn my palm, add a fireball
to my hand}''
and have these effects trigger automatically, without masks, temporal annotations, 
or per-clip scripting that requires manual or extra annotations, as demonstrated in the bottom row of \cref{fig:teaser}. 
Beyond reacting to a single event, a system should also distinguish which of several specified events occurs and apply the corresponding edit, i.e., support multi-event-triggered editing (see \cref{fig:comp_figure2}).

We study \textbf{event-triggered egocentric video editing}: generative, trigger-conditioned editing from an egocentric RGB stream. Given a monocular video and a natural-language rule of the form ``\emph{if $X$ happens, apply edit $Y$},'' a model must detect which event, if any, occurs and apply the corresponding edit only to subsequent frames, leaving the pre-event stream untouched, as shown in \cref{fig:teaser}.
We also consider multi-event prompts such as ``\emph{if $X_1$ happens, apply $Y_1$; if $X_2$ happens, apply $Y_2$}.''
Event-triggered video editing is challenging due to the need for \emph{joint event understanding and editing} without boundary artifacts, 
\emph{temporal restraint} on non-occurring triggers, \emph{causal inference} without access to future context, and \emph{compositional prompt-following} under multiple events.

Existing approaches address only fragments of this problem. Instruction-based editors~\cite{kulikov2024flowedit,InstructVid2Vid,lucyedit2024} typically apply a single edit persistently across the input or user-specified regions, and do not condition the edit on a temporally \textit{localized event}. 
Mask- and reference-based methods~\cite{wananimate2025,ye2025unic,editverse2025,mai2025easyv2v} provide stronger localized control, yet require manual inputs such as masks, 
making them unsuitable for autonomous streaming deployment. 
Moreover, these methods do not support event-triggered editing. 
Event detection pipelines~\cite{shou2021generic, wang2023temporal} can localize actions but are decoupled from editing, introducing a delay between the detected trigger and the applied edit.
Critically, to our knowledge, no prior system provides training data, training recipes, or evaluation protocols 
specifically designed for \emph{event-triggered video editing}. 
This requires a joint capacity for event understanding and temporally grounded editing, 
which remains absent even in bidirectional settings, 
where the model can access future frames, and is substantially more challenging in causal or streaming mode.

\revision{We introduce \textbf{EgoPlay}, an end-to-end event-triggered V2V editor obtained by fine-tuning a pretrained EgoEdit video diffusion transformer. Our contribution is the task, data, training recipe, causal variant, and evaluation protocol rather than a new foundation architecture. The training set is built primarily from public Ego4D footage, with a smaller human-labeled general-domain subset for diversity. EgoPlay is built on three key ideas.}
First, we formulate trigger-conditioned V2V as a unified task, 
enabling joint learning of event recognition and pixel-level editing so that edit decisions depend on the visual events observed in the video.
Second, we construct a large-scale event-triggered editing dataset using a multi-stage VLM pipeline, generating positive-trigger, 
fabricated-trigger negative, and multi-event prompts together with edited video pairs.
Third, we train a bidirectional event-triggered video editor with multi-event prompt supervision and 
derive a \emph{causal} variant with diffusion forcing for chunk-by-chunk deployment.
While our bidirectional model serves as an offline upper bound, the streamable causal variant nearly matches its editing quality despite operating chunk-by-chunk without future context.

To evaluate this new event-triggered V2V capability, 
we introduce an event-aware protocol that splits each sequence at the ground-truth event boundary and independently scores pre-event preservation, 
post-event editing quality, 
and robustness to fabricated triggers.
On the Ego4D benchmark, EgoPlay achieves the best averaged event-triggered performance among methods without ground-truth event boundaries, outperforming both a strong instruction-based editor and an explicit detector-then-edit cascade, 
with especially clear gains on positive and compositional event-triggered modes, producing edits that are largely confined to the post-event segment.

\revision{A modular detector--editor is a natural alternative, and we evaluate it explicitly. It can reject absent events by copying the input, but detection errors become hard temporal cuts: the editor cannot repair a missed or misplaced boundary after the prefix and suffix have been stitched. It also requires a resident VLM and editor. The unified model instead learns a soft visual transition in one model and can be converted into a causal student. This design targets always-on wearable capture, where a user specifies a persistent rule once and the system reacts to first-person actions without a mask, timestamp, or per-clip script; we do not claim that modular agents are universally inferior.}

\paragraph{Contributions}
\begin{itemize}
    \item We formulate \textbf{event-triggered video editing} as a V2V task in which a model must detect visual triggers from raw pixels, preserve pre-trigger frames, and apply the requested edit only after the trigger.
    \item We construct a large-scale \textbf{event-triggered editing dataset} with 106K pairs spanning positive, fabricated-negative, and multi-event prompts with edited targets.
    \item We propose \textbf{EgoPlay} in two variants: a bidirectional model that provides an offline upper bound for event-triggered editing, and a causal model for streamable inference.
    \item We introduce an \textbf{event-aware evaluation protocol} that 
    separately measures pre-event restraint, post-event editing quality, 
    and false-trigger robustness. EgoPlay achieves the best averaged event-triggered performance among methods that do not receive ground-truth event boundaries.
\end{itemize}

%% file: figures/teaser.tex
\begin{figure}
    \centering
    \includegraphics[width=\columnwidth]{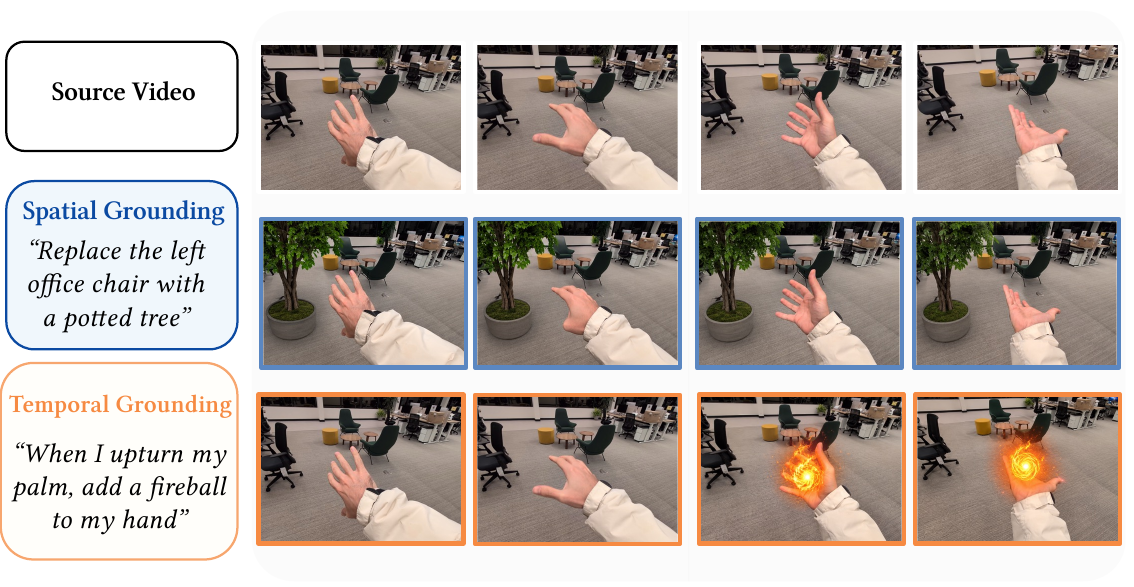}
    \caption{
    \textbf{Event-Triggered video editing.}
    Given a source video and a natural-language instruction, a conventional video editor can ground \emph{where} to edit, but always applies the edit throughout the clip. In contrast, EgoPlay grounds \emph{when} to edit from an event-triggered prompt: it preserves frames before the specified event and applies the edit only after the event occurs.
    }
    \Description{
    Teaser figure comparing a source video, a normal spatially grounded video edit, and EgoPlay's temporally grounded event-triggered edit.
    }
    \label{fig:teaser}
\end{figure}

%% file: sections/2_related.tex
\section{Related Work}
\input{figures/data_pipe}

\paragraph{Egocentric World Models.}
Recent egocentric world models move beyond passive first-person video understanding toward controllable simulation of embodied experience. 
Several systems condition generation on human or agent actions: 
PlayerOne~\cite{tu2025playerone} synthesizes egocentric video from body motion with disentangled motion injection, 
EgoWM~\cite{bagchi2026egowm} adapts pretrained video diffusion models into action-conditioned world models through lightweight timestep modulation, 
and LOME~\cite{gao2026lome} jointly predicts video and spatial action maps for fine-grained hand--object manipulation. 
Related work also connects first-person perception to interactive XR and 3D reconstruction. 
Generated Reality~\cite{xie2026generatedreality} uses VR-tracked hand and head poses for real-time human-centric world simulation, 
EgoReAct~\cite{zhang2025egoreact} predicts 3D human reaction motions from streaming egocentric video, 
and FunRec~\cite{delitzas2026funrec} reconstructs functional articulated 3D scenes from egocentric interaction videos. 
These works demonstrate the value of first-person visual dynamics as a generative prior. In contrast, EgoPlay studies a complementary problem: 
editing an egocentric stream only after a user-specified visual event occurs, while preserving all pre-event content.

\paragraph{Streaming Video Understanding.}
Streaming video understanding focuses on maintaining useful state over long or continuous visual input under causal constraints. 
Recent methods improve memory management and online reasoning with compact caches or persistent event memory~\cite{zhang2026hermes,xiong2025streamchat,wang2025streamforest}. 
StreamingVLM~\cite{xu2025streamingvlm} enables real-time understanding of unbounded videos with attention sinks and sliding-window KV caches, 
while Streamo~\cite{xia2025streamo} builds instruction-tuning data for streaming tasks such as narration, 
temporal grounding, and time-sensitive question answering. 
StreamMind~\cite{ding2025streammind} further introduces event-gated cognition, 
invoking expensive language-model reasoning only when relevant events occur. 
EgoPlay shares the causal and event-aware setting with these systems, 
but targets generative video editing rather than recognition or dialogue: 
the model must not only detect that an event has happened,
but also synthesize a temporally coherent edited continuation from that point onward.

\paragraph{Video Editing.}
Instruction-guided editing has progressed rapidly for images, 
where paired synthetic supervision and increasingly capable editors such as InstructPix2Pix~\cite{brooks2023instructpix2pix}, 
UltraEdit~\cite{zhao2024ultraedit}, and GPT-4o-style image editors~\cite{openai2024gpt4ocard} 
show that natural-language edit instructions can provide strong controllability. 
Extending these ideas to video is harder because the edit must remain temporally consistent while preserving identity, 
motion, and background structure. Early and training-free video editors, 
including VidEdit and AnyV2V~\cite{couairon2024videdit,ku2024anyv2v}, 
edit a given clip without large paired V2V training. 
Later approaches adapt image editors to video or train directly on synthetic instruction-based video pairs, 
including EVE~\cite{singer2024eve}, InstructVid2Vid~\cite{InstructVid2Vid}, 
Consistent Video-to-Video Transfer~\cite{cheng2023consistentvideotovideotransferusing}, 
EffiVED~\cite{zhang2024effived}, and VEGGIE~\cite{yu2025veggie}. More recent systems scale data, 
architectures, or in-context conditioning for stronger offline video editing, 
such as LucyEdit~\cite{lucyedit2024}, EditVerse~\cite{editverse2025}, VACE~\cite{vace}, Runway Aleph~\cite{runway2025aleph}, 
EasyV2V~\cite{mai2025easyv2v}, and in-context video editing with unpaired clips~\cite{icve}.
A separate line of work targets \emph{streaming} editing and generation, 
converting bidirectional diffusion teachers into causal, chunk-by-chunk students for online inference. 
A series of autoregressive training paradigms drives this transition, including teacher forcing~\cite{jin2025pyramidalflow},diffusion forcing~\cite{chen2024diffusionforcing}, self forcing~\cite{huang2025selfforcing}, 
and causal forcing~\cite{zhu2026causalforcing}, often paired with distribution-matching distillation~\cite{yin2024dmd, yin2024dmd2} that compresses a bidirectional teacher into a few-step causal student, as in ~\cite{yin2025causvid,huang2025selfforcing,zhu2026causalforcing}.
Most relevant to us, EgoEdit~\cite{egoedit}, paves a way for egocentric streaming video editing. 

\revision{Recent instruction editors Ditto~\cite{bai2025ditto} and VIVA~\cite{cong2026viva} provide strong event-free V2V baselines, while commercial systems such as Runway Aleph~\cite{runway2025aleph} combine language-model orchestration, visual understanding, image/video editing, and stitching. Such modular agents can use specialized detectors and editors, but their event boundary is an explicit handoff: localization errors propagate to the generated suffix and the join can introduce a visible discontinuity. EgoPlay studies whether event localization and editing can instead be learned jointly inside one editor, and compares both designs under the same event-aware protocol.}

Despite this progress, existing editors generally assume an offline edit request that is applied to the entire input, 
a manually specified temporal span, or an explicitly localized region. 
Mask-, reference-, and in-context-based methods can provide stronger spatial 
or identity control~\cite{wananimate2025,ye2025unic,omniinsert2025,editverse2025}, 
but they still require user-provided control signals or operate after the target clip is available. 
EgoPlay instead addresses \emph{event-triggered} editing in a causal stream: the model receives no event timestamp or mask at inference time, 
must infer when the trigger occurs from visual evidence, leave pre-event frames unchanged, and generate the edited continuation online after the trigger.

%% file: figures/data_pipe.tex
\begin{figure*}[!t]
  \centering
   \includegraphics[width=\linewidth]{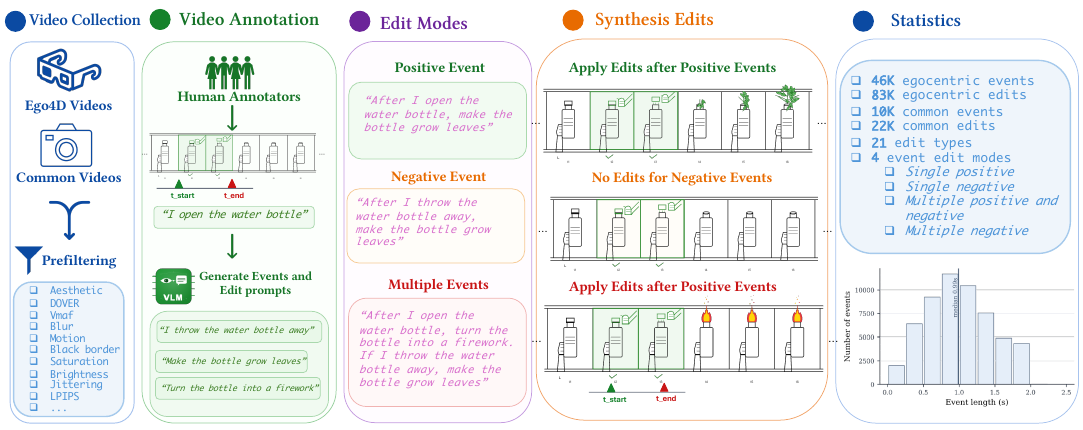}
   \vspace{-2em}
   \caption{\textbf{Data pipeline.} Given raw videos and human-annotated event spans, we (1) filter low-quality clips and curate events with clear completion points, (2) generate event-triggered prompts via a multi-pass VLM that produces positive, negative, and event-free instructions, and (3) synthesize edited targets by applying an event-free edit to the post-event segment and blending with VAE transition frames to achieve a smooth editing transition without an obvious temporal boundary. The resulting training set is built primarily from Ego4D (81,682 egocentric videos), with a smaller auxiliary common-video subset (21,961 videos) included for diversity.}
   \label{fig:data_pipe}
\end{figure*}

%% file: sections/3_method.tex
\section{Method}

EgoPlay learns to jointly detect when a user-specified event occurs
and to apply the requested edit from that moment onward. %
We formalize each instruction as a prompt ``if $X$ then $Y$,''
where $X$ is a visual event and $Y$ is an edit, and allow multi-event
prompts that concatenate several event--edit pairs.
This differs from conventional streamable video generation, which
typically uses a fixed prompt over the entire causal trajectory and
is not triggered by visual events.
Training such a model requires examples containing the source video,
the edited target video, the conditional prompt, and the temporal
localization of the event/edit boundary, together with negative-event
examples where the condition $X$ is not satisfied and the target remains
unchanged.
We first describe how we build the training data~(\S\ref{sec:data}),
then present the bidirectional model~(\S\ref{sec:bidir})
and the causal variant derived from it~(\S\ref{sec:causal}).

\subsection{Event-Triggered Data Generation}
\label{sec:data}

We build event-triggered editing data primarily from the public Ego4D~\cite{ego4d} egocentric video dataset, 
with a smaller auxiliary common-video subset used only to increase visual and event diversity during training.
For both sources, we retain visually grounded events that can serve as reliable triggers for event-triggered video editing.
\cref{fig:data_pipe} summarizes our data pipeline.

\paragraph{Ego4D~\cite{ego4d}.}
Ego4D is substantially noisier because it is captured from
head-mounted first-person cameras, so we apply stronger automatic filtering.
We first remove low-quality clips via video-quality filters covering aesthetics, blur, motion, brightness, saturation, jitter, and related artifacts.
A vision-language model (VLM)~\cite{bai2025qwen3} then retains only narrations that describe discrete, 
visually observable actions with a clear completion point, discarding prolonged states or ambiguous activities.
The VLM further verifies narration–visual alignment and rewrites the retained event into clean first-person text.
\revision{The exact thresholds and filtering prompts are provided in the supplement; only high-confidence visual matches are retained.}

\paragraph{Auxiliary common-video subset.}
\revision{We additionally include a smaller set of human-labeled general-domain, predominantly third-person videos for visual and event diversity; this subset is not presented as egocentric.}
Annotators provide event spans and descriptions; we retain events that are within the training clip timeframe, 
visually identifiable, and have a clear completion point, with sufficient context before and after the trigger.
We rebalance event locations within the sampled window to diversify the temporal distribution of triggers.

\paragraph{Prompt generation.}
After event curation, we run a multi-pass VLM pipeline to generate event-triggered prompts.
We sample a post-event frame for the VLM to identify which edit types from our catalog of 21 categories are visually feasible.
For four feasible edit types, the VLM first generates event-free
instructions used for video synthesis.
The VLM then generates positive event-triggered prompts grounded in the
post-event frame, the event text, and edit-type-specific guidance.
We rewrite these prompts into diverse conditional templates and separately strip the trigger clause 
to produce event-free instructions used during video synthesis.
For \emph{negatives}, we compose each sample with multiple diverse but non-occurring events in the same scene, 
paired with diverse edit instructions as well.
This yields paired positive and negative event-triggered prompts, 
directly supervising the model on \emph{when} editing should occur and when it should not.
\revision{For example, the retained event ``I pick a rasp from the floor'' yields the positive prompt ``After I pick up the rasp, add a woven leather wristband to my left wrist'' and the event-free synthesis prompt ``Add a woven leather wristband to the left wrist.'' A hard negative replaces the trigger with a plausible absent action, e.g., ``After I drop the rasp onto the tube, add the wristband.'' Additional examples and the generation templates are given in the supplement.}

\paragraph{Event-Triggered V2V data synthesis.}
\revision{Given a positive prompt, we run the pretrained EgoEdit model on the post-event segment using the corresponding event-free instruction, yielding edited frames}
$g$.
At the event boundary, a hard cut from source to edited frames would create an abrupt change. We therefore insert a short transition: \revision{using the VAE from Qwen-Image-2512~\cite{qwenimage2025}, we spherically interpolate corresponding source and edited latents and decode} $M$ frames $h_{e+n_{evt}},\ldots,h_{e+n_{evt}+M-1}$ that blend the source appearance into the edited appearance over $M$ frames.
Let $n_{pre},n_{evt},n_{post}$ denote the frame counts before the event $e$,
during the event, and after the event; $f$ denotes source frames.
The source clip and positive training target are:

{\small
\begin{align}
S ={}& \underbrace{[f_{e-n_{pre}},\ldots,f_{e-1}]}_{\text{pre-event}}
       +\underbrace{[f_{e},\ldots,f_{e+n_{evt}-1}]}_{\text{event}} \notag\\
     &+\underbrace{[f_{e+n_{evt}},\ldots,f_{e+n_{evt}+n_{post}-1}]}_{\text{post-event}}, \\[6pt]
T^{+} ={}& \underbrace{[f_{e-n_{pre}},\ldots,f_{e+n_{evt}-1}]}_{\text{preserve pre-event}}
         +\underbrace{[h_{e+n_{evt}},\ldots,h_{e+n_{evt}+M-1}]}_{\text{transition}} \notag\\
       &+\underbrace{[g_{e+n_{evt}+M},\ldots,g_{e+n_{evt}+n_{post}-1}]}_{\text{edited post-event}}.
\end{align}
}
For a negative prompt the target is unchanged: $T^{-}=S$.

\paragraph{Multi-event prompt training.}
During training we also sample \emph{multi-event} prompts by
concatenating one anchor prompt (positive or negative) with one
to three additional negative prompts in random order.
This compositional setting teaches the model to selectively trigger
on the positive event while ignoring distractors — the setting where our approach shows its largest gains (\cref{tab:main_results}).

\paragraph{Dataset scale.}
After quality filtering and temporal rebalancing, the final dataset contains 81{,}682 Ego4D training pairs
and 21{,}961 auxiliary common-video training pairs, plus 2{,}408 held-out test pairs (1{,}897 Ego4D and 511 auxiliary), for a total of \textbf{106{,}051} event-triggered clip--prompt pairs.

\input{figures/method}

\input{tables/main_results}

\input{figures/comp_figure}
\subsection{Bidirectional Event-Triggered Editor}
\label{sec:bidir}

The central challenge in event-triggered V2V is that the model
must solve two tightly coupled problems simultaneously: it must
\emph{detect} when the specified event occurs directly from raw pixels, and \emph{apply} the requested edit persistently from that
moment onward — all within a single forward pass, with no separate
event detector, no manual boundary annotation, and no post-hoc
stitching, which would introduce latency and break temporal
consistency.
We achieve this by conditioning a diffusion transformer~\cite{Peebles2023DiT} on both the source video and the event-triggered instruction end-to-end, so that event localization and pixel-level editing emerge as a
unified learned behavior.

\paragraph{Architecture.}
We adapt a pretrained text-to-video diffusion transformer (DiT)~\cite{Peebles2023DiT}
to a video-to-video formulation.
The source clip $S$ and target clip $T$ are encoded by a frozen
video VAE.
We condition the model on the source latent via \emph{channel-wise concatenation} with the noisy target: 
source tokens and noisy target tokens are produced by two parallel patch embeddings and
summed at the transformer input, introducing source conditioning at zero cost in sequence length while preserving the pretrained backbone. 
The event-triggered instruction is injected through cross-attention, 
binding the \emph{when} (the trigger clause) and the \emph{how} (the edit specification) into a single conditioning signal.

Crucially, the model never receives an explicit event timestamp as input, at either training or inference time. During training, the boundary is encoded only in how the target $T^{+}$ is constructed; at inference, the model must infer the trigger boundary from the visual content of the source clip itself and apply the edit only to frames after the event, learning implicit event localization as a byproduct of the editing objective. This is what distinguishes event-triggered video editing from temporal-mask editing: a temporal mask supplies the boundary explicitly as input, whereas here the model must localize it from pixels alone and perform event location and editing end-to-end.

\paragraph{Training objective.}
We train with rectified flow matching~\cite{lipman2022flow}.
Let $z_0$ be the clean target latent,
$\epsilon \sim \mathcal{N}(0,\sigma^2 I)$,
and $t \in [0,1]$ sampled from a logit-normal schedule.
We form the noisy latent $z_t = (1-t)\,z_0 + t\,\epsilon$
and define the velocity target $v^{\star} = \epsilon - z_0$.
We collect the source latent, text embedding, and auxiliary
metadata into a single conditioning signal $c$, and train
$v_\theta$ with:
\begin{equation}
  \mathcal{L}_{\mathrm{RF}}
  = \mathbb{E}_{z_0,\epsilon,t}\!
    \left[\,
      \bigl\|v_\theta(z_t,\,c,\,t) - v^{\star}\bigr\|_2^2
    \right].
\end{equation}
The supervision signal is entirely pixel-level: the model is
never told \emph{where} the event boundary is during training.
Instead, it must discover the boundary implicitly in order to
produce the correct target $T^+$ — editing post-event frames
while leaving pre-event frames identical to the source.
Positive, negative, and multi-event prompts are sampled so the model
simultaneously learns when editing should occur and when the source
must be preserved exactly, i.e., when the event is never triggered.
The result is a single model that detects, localizes, and
edits in one pass, without any cascade, and without any explicit localization module
(see Fig.~\ref{fig:method} left).

\subsection{Causal Event-Triggered Editor}
\label{sec:causal}

The bidirectional editor attends to all frames jointly, which makes it
a strong offline upper bound but prevents streaming deployment.
We derive a streaming-compatible variant by keeping the same
event-triggered V2V formulation and replacing visual self-attention
with blockwise causal self-attention.
The latent sequence is divided into temporal blocks
\(\{B_1,\ldots,B_N\}\).
Tokens inside a block attend bidirectionally within that block, while
tokens in block \(B_i\) can attend only to blocks \(B_{\leq i}\) and
never to future blocks \(B_{>i}\).
The text cross-attention remains unchanged because the instruction is
available globally, while source-video conditioning is injected
framewise before the transformer and is therefore constrained by the
same visual self-attention mask.

We train this causal model (See Fig.~\ref{fig:method}) with diffusion forcing~\cite{chen2024diffusionforcing}.
For each temporal block \(B_i\), we sample an independent noise level
\(t_i\) and apply the rectified-flow noising process blockwise: $z_{\mathbf{t}}^{(i)} = (1-t_i)\,z_0^{(i)} + t_i\,\epsilon^{(i)}$ for $i=1,\ldots,N$.
The model receives the vector of per-block noise levels
\(\mathbf{t}=(t_1,\ldots,t_N)\), repeated over frames within each
block, and predicts the same velocity target
\(v^\star=\epsilon-z_0\):
\begin{equation}
  \mathcal{L}_{\mathrm{DF}}
  =
  \mathbb{E}_{z_0,\epsilon,\mathbf{t}}
  \left[
    \sum_{i=1}^{N}
    \bigl\|
      v_\theta(z_{\mathbf{t}},c,\mathbf{t})^{(i)}
      - v^{\star,(i)}
    \bigr\|_2^2
  \right].
\end{equation}
This matches streaming inference: when generating block \(B_i\), the previous blocks have already been denoised and stored in the KV cache,
while the current block is still noisy.
The bidirectional model, therefore, establishes the task upper bound, whereas the causal variant performs chunk-by-chunk editing without
future-block access.

%% file: figures/method.tex
\begin{figure}
    \centering
    \includegraphics[width=\columnwidth]{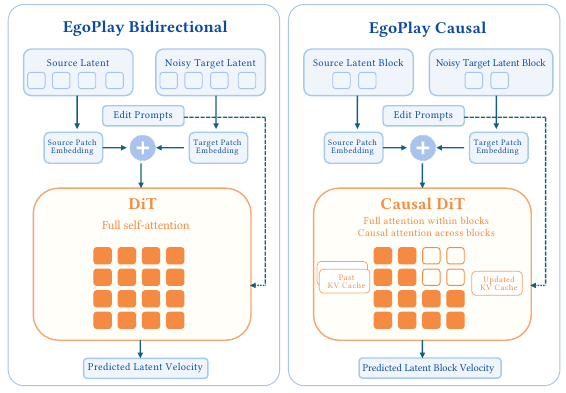}
    \vspace{-2em}
    \caption{\textbf{EgoPlay method.} EgoPlay takes a source video and an event-triggered instruction, jointly infers when the trigger event occurs and what edits to apply with the event. The bidirectional model learns this behavior end-to-end from event-triggered V2V supervision, while the causal variant restricts temporal attention for streamable inference.}
    \label{fig:method}
\end{figure}

%% file: tables/main_results.tex
\begin{table}[t]
\centering
\scriptsize
\caption{Main event-triggered editing results on the Ego4D benchmark. We evaluate 150 source videos for each reported event-triggered mode, yielding 600 event-mode samples in total. For each mode, we report editing quality, visual quality, background consistency, and total score. The final block averages over the four event-triggered modes. Oracle GT target uses the ground-truth event boundary with the EgoEdit model to edit the post-event segment. Higher is better.}
\label{tab:main_results}
\vspace{-2em}
\resizebox{\columnwidth}{!}{%
\begin{tabular}{llcccc}
\toprule
Mode & Method & \shortstack{Editing\\quality} & \shortstack{Visual\\quality} & \shortstack{Background\\consistency} & Total \\
\midrule
\shortstack[l]{Single\\positive}
& EgoEdit & 2.20 & 2.21 & 2.23 & 6.64 \\
& VLM-guided & 1.64 & 1.70 & 1.76 & 5.09 \\
& EgoPlay & \textbf{2.67} & \textbf{2.67} & \textbf{2.68} & \textbf{8.02} \\
& EgoPlay-Causal & 2.59 & 2.61 & 2.61 & 7.81 \\
& \textcolor{gray}{Oracle GT target} & \textcolor{gray}{2.79} & \textcolor{gray}{2.79} & \textcolor{gray}{2.81} & \textcolor{gray}{8.39} \\
\midrule
\shortstack[l]{Multiple positive\\and negative}
& EgoEdit & 1.91 & 1.95 & 2.00 & 5.86 \\
& VLM-guided & 1.58 & 1.62 & 1.68 & 4.88 \\
& EgoPlay & \textbf{2.49} & \textbf{2.51} & \textbf{2.52} & \textbf{7.52} \\
& EgoPlay-Causal & 2.46 & 2.48 & 2.48 & 7.42 \\
& \textcolor{gray}{Oracle GT target} & \textcolor{gray}{2.80} & \textcolor{gray}{2.79} & \textcolor{gray}{2.80} & \textcolor{gray}{8.39} \\
\midrule
\shortstack[l]{Single\\negative}
& EgoEdit & 2.54 & 2.54 & 2.54 & 7.63 \\
& VLM-guided & \textbf{3.00} & \textbf{2.99} & \textbf{3.00} & \textbf{8.99} \\
& EgoPlay & 2.78 & 2.77 & 2.78 & 8.33 \\
& EgoPlay-Causal & 2.75 & 2.74 & 2.74 & 8.22 \\
& \textcolor{gray}{Oracle GT target} & \textcolor{gray}{3.00} & \textcolor{gray}{2.99} & \textcolor{gray}{3.00} & \textcolor{gray}{8.99} \\
\midrule
\shortstack[l]{Multiple\\negative}
& EgoEdit & 2.40 & 2.42 & 2.42 & 7.24 \\
& VLM-guided & \textbf{3.00} & \textbf{2.99} & \textbf{3.00} & \textbf{8.99} \\
& EgoPlay & 2.72 & 2.71 & 2.72 & 8.14 \\
& EgoPlay-Causal & 2.67 & 2.66 & 2.67 & 8.00 \\
& \textcolor{gray}{Oracle GT target} & \textcolor{gray}{3.00} & \textcolor{gray}{2.99} & \textcolor{gray}{3.00} & \textcolor{gray}{8.99} \\
\midrule
\shortstack[l]{Average\\over modes}
& EgoEdit & 2.26 & 2.28 & 2.30 & 6.84 \\
& VLM-guided & 2.30 & 2.33 & 2.36 & 6.99 \\
& EgoPlay & \textbf{2.66} & \textbf{2.67} & \textbf{2.68} & \textbf{8.01} \\
& EgoPlay-Causal & 2.62 & 2.62 & 2.63 & 7.86 \\
& \textcolor{gray}{Oracle GT target} & \textcolor{gray}{2.90} & \textcolor{gray}{2.89} & \textcolor{gray}{2.90} & \textcolor{gray}{8.69} \\
\bottomrule
\end{tabular}
}
\end{table}

%% file: figures/comp_figure.tex
\begin{figure*}[t]
  \centering
   \includegraphics[width=\linewidth]{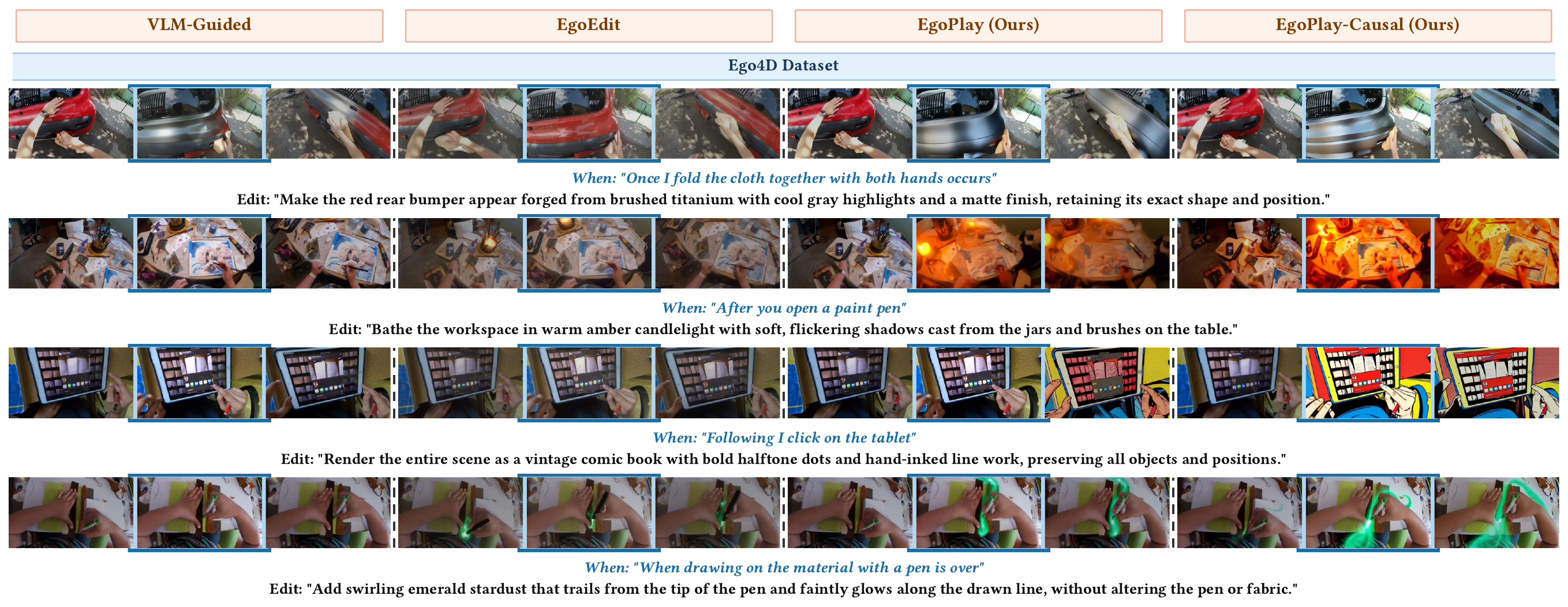}
      \vspace{-2em}
   \caption{\textbf{Qualitative comparison on Ego4D event-triggered video editing.} We compare EgoPlay and EgoPlay-Causal against VLM-guided editing and EgoEdit in the single-positive setting. For each example, three frames are shown: before, during (highlighted), and after the edit. Each edit is conditioned on a temporal trigger (When, italic) specifying the moment the edit should take effect, along with a spatial instruction (Edit, bold). EgoPlay better respects the trigger timing while maintaining edit fidelity and temporal coherence in egocentric scenes.}
   \Description{Qualitative Ego4D dataset comparison for event-triggered video editing across EgoPlay, EgoPlay-Causal, VLM-guided editing, and EgoEdit.}
   \label{fig:comp_figure}
\end{figure*}

%% file: sections/4_experiment.tex
\section{Experiments}

\input{tables/editversebench}

\subsection{Evaluation Protocol}

We evaluate EgoPlay on the curated Ego4D test split produced by our data pipeline. 
The evaluation is designed to test both parts of the event-triggered editing problem: 
whether a method recognizes that the requested event has occurred, and whether it applies the edit only after the event while preserving all irrelevant content.
\revision{The selected benchmark contains \benchmarksourcecount{} event--prompt pairs evaluated under five prompt modes. In the main table (Table~\ref{tab:main_results}) we report the four event-triggered modes, totaling \benchmarkreportedsamples{} event-mode samples. The fifth mode is standard event-free V2V editing.}

\paragraph{Compared methods.}
We compare our EgoPlay to two baselines.
\textbf{EgoEdit}\egoeditcite{} is the state-of-the-art instruction-guided V2V editor for egocentric videos; it receives an event-free edit instruction and edits the input clip directly. \revision{We obtained model access from the EgoEdit authors and ran it ourselves on this benchmark.}
It measures how far a strong standard editor can go without event-triggered reasoning. 
\textbf{VLM-guided editing} factorizes the task into explicit event understanding followed by standard editing. 
We use Qwen3-VL-8B~\cite{bai2025qwen3}, which provides a practical balance between efficiency and performance.
A VLM reads the source clip and event-triggered prompt, predicts whether the trigger is present, estimates the trigger end time, and rewrites the request into an event-free instruction. 
If the event is detected, an EgoEdit editor is applied to the predicted post-event suffix and the result is stitched back to the unedited prefix; otherwise, the source video is copied.
Our \textbf{EgoPlay} processes the full 81-frame source clip and the original event-triggered prompt end-to-end, performing joint event recognition and editing. 
It receives no event timestamp at inference time and must infer from visual evidence whether and when the edit should begin.
\ifsubmittedbenchmark\else
We also report an \textbf{Oracle GT target} upper bound that uses the ground-truth event boundary with the EgoEdit model, editing only the post-event segment and preserving the preceding frames.
\fi

\paragraph{VLM-based evaluator.}
We score each predicted video with an independent VLM evaluator under three criteria: editing quality, visual quality, and background consistency~\cite{editverse2025}.
The evaluator is intentionally blind to the event-triggered prompt. 
It never receives the trigger clause, fabricated negative events, or other event-conditioning text used at inference time. 
Instead, it compares aligned source--prediction clip pairs. 
For pairs that should be edited, the evaluator receives the canonical event-free edit instruction, the source clip, and the predicted clip. 
For pairs that should remain unchanged, it receives no edit instruction and judges whether the prediction correctly preserves the source. 
\revision{The evaluator is Qwen3-VL-30B-A3B, a separate and larger model than the 8B detector used by the modular baseline.} This design prevents the evaluator from rewarding textual agreement with the trigger clause and forces event-timing errors to appear as visible video errors. \revision{Agreement with human preferences and independently annotated edit-start accuracy provides two additional signals; nevertheless, we report known evaluator failures in the supplement.}

\paragraph{Temporal decomposition.}
Let $S$ denote the source clip and $G$ the predicted clip. 
Using the ground-truth trigger boundary, we decompose both clips into a pre-event segment of length $n_{pre}$, 
an event segment of length $n_{evt}$, and a post-event segment of length $n_{post}$. We use $f$ for source frames and $g$ for predicted frames, 
and form three aligned source--prediction pairs:
\begin{align}
P_{n_{pre}} &= ([f_{e-n_{pre}}, \ldots, f_{e-1}], [g_{e-n_{pre}}, \ldots, g_{e-1}]), \\
P_{n_{evt}} &= ([f_e, \ldots, f_{e+n_{evt}-1}], [g_e, \ldots, g_{e+n_{evt}-1}]), \\
P_{n_{post}} &= ([f_{e+n_{evt}}, \ldots, f_{e+n_{evt}+n_{post}-1}], [g_{e+n_{evt}}, \ldots, g_{e+n_{evt}+n_{post}-1}]).
\end{align}
For a positive sample, the desired target is $T^+$: $P_{n_{pre}}$ and $P_{n_{evt}}$ should remain unchanged, while $P_{n_{post}}$ should contain the requested edit. 
For a negative sample, the desired target is $T^- = S$, so all three segments should remain unchanged. 
Multi-event prompts follow the same condition: a sample is positive only if at least one true trigger event is present, and negative otherwise.

\paragraph{Fair scoring.}
Visual quality measures realism and artifact level, while background consistency measures whether regions unrelated to edit remain stable. 
Both are scored on the same temporal segments and aggregated with the same weights as editing quality. 
All methods are evaluated using the same event-free evaluation instruction with ground-truth temporal boundary. 
For the VLM-guided baseline, the VLM-predicted boundary is used only to generate its output, not to score it. 
Thus, localization mistakes are naturally penalized as edits that occur too early, too late, or not at all.

\subsection{Event-Triggered Editing Results}

Table~\ref{tab:main_results} reports the main event-triggered editing results on Ego4D. 
We evaluate four modes: single positive, multiple positive and negative, single negative, and multiple negative. 
These modes test complementary capabilities: positive modes require the model to detect the trigger and edit after it, 
while negative modes require the model to reject plausible but absent triggers and preserve the source.
Among methods that do not receive the ground-truth event boundary, 
EgoPlay achieves the best averaged event-triggered performance, 
with the highest average editing quality, visual quality, background consistency, and total score over the four modes. 
The improvement is most pronounced in positive modes, 
where event recognition and editing must be solved jointly: 
EgoPlay leads both the single-positive and multiple-positive-and-negative settings 
over EgoEdit and the VLM-guided detector--editor cascade (See Fig.~\ref{fig:comp_figure} and Fig.~\ref{fig:comp_figure2}). 
\ifsubmittedbenchmark\else
The Oracle GT target row remains higher overall, as expected,
because it uses ground-truth temporal boundaries and therefore measures the remaining gap to perfect event localization.
\fi

\revision{To broaden the comparison beyond egocentric editors, we additionally evaluate the recent open-source video editors Ditto and VIVA on the submitted 40-pair benchmark (\additionalbaselinesref). Their four-mode totals are 4.72 and 5.77, compared with EgoPlay's 7.72. Because these editors have no trigger representation, they apply the event-free edit even when the condition is absent. A complementary five-clip commercial-system comparison (\runwayresultsref) finds that Runway Aleph2's agent obtains a higher VLM score but introduces larger boundary discontinuities, while Runway V2V scores below EgoPlay-Causal. \baselineseparationtext}

The negative modes reveal a complementary trade-off. 
The VLM-guided cascade performs best among non-oracle methods when no trigger is present 
because it can choose to copy the source video once the detector rejects the event. 
EgoPlay also preserves many no-edit cases, but its end-to-end generative formulation is less conservative than the explicit detector--copy baseline. Nevertheless, EgoPlay's positive-trigger gains outweigh this gap in the averaged Ego4D benchmark, yielding the strongest overall event-triggered result among methods without ground-truth boundaries.

\subsection{Human Preference Study}

\revision{We conducted a human preference study on one fixed, non-hand-picked set of 30 positive samples (five source videos, each with six distinct edits). All ten participants evaluated every sample. Each trial showed the ground-truth target and the four methods' outputs in anonymized, randomized order; the target uses the human-annotated boundary, preserves the pre-event and event frames, and applies the edit only afterward. For each baseline, this gives $30\times10=300$ EgoPlay--baseline judgments per criterion. Participants indicated a preference---or tie---on four criteria:}
instruction following, edit timing, edit quality, and overall preference. EgoPlay is preferred over all baselines on overall preference and on the large majority of criteria; Table~\ref{tab:userstudy} gives the per-question breakdown, and the study interface is reproduced in the supplementary material. Aggregated across criteria, EgoPlay's win rate against every baseline significantly exceeds the 50\% chance level (Wilson 95\% CI); the only per-criterion exception is edit quality against the near-equivalent EgoPlay-Causal variant, where the two methods tie.
The smallest margin is against EgoPlay-Causal, confirming that in perceived editing quality the causal variant reaches close to the bidirectional upper bound.
\input{tables/table_user_study}

\subsection{Standard Video Editing}
A desirable event-triggered framework should retain the ability to perform ordinary instruction-guided video editing when no trigger clause is required. We therefore compare the underlying editing models on EgoEditBench and EditVerseBench, two recent V2V editing benchmarks. The comparison covers attention-manipulation methods, first-frame propagation methods, instruction-guided editors, and streaming variants. Table~\ref{tab:egoedit_comparison} shows that the standard EgoEdit backbone is competitive with or stronger than prior offline editors across both benchmarks, while EgoEdit-RT provides a strong streaming counterpart. These results indicate that the editing model used by EgoPlay starts from a capable conventional V2V editor. The event-triggered training studied in Table~\ref{tab:main_results} then adds trigger-aware preservation and post-event editing behavior, rather than replacing the standard editing capability with a separate detector or manual temporal mask.

\subsection{Ablation Study}

We analyze two diagnostic factors that are central to event-triggered editing: whether the visible edit begins near the true event boundary, and whether the event-triggered formulation changes inference cost.

\paragraph{Edit-timing ablation.}
A central claim of EgoPlay is that the model implicitly localizes when the trigger event ends and starts editing afterward. To measure this behavior directly, we ask human annotators to mark the first frame where an edit becomes visible in each predicted video. Let $\hat{e}$ denote this predicted edit-start frame, $e^{\star}$ the ground-truth trigger-end frame, and $T$ the total number of frames. We report event-end localization accuracy at tolerance ratio $r$ as $\mathrm{Acc}@r = \mathbf{1}[|\hat{e} - e^{\star}| \leq rT]$, $r \in \{0.1, 0.2, 0.3\}$.
We restrict this diagnostic to positive event-triggered samples, since negative samples should contain no edit start.
For the VLM-guided baseline, we use the event-end frame predicted by the VLM directly.

\input{tables/event_timing}

Table~\ref{tab:event_timing} shows that EgoPlay gives the most precise edit-start alignment at the strictest tolerance, improving Acc@0.1 from 40.0 to 45.5. At more relaxed thresholds, the VLM-guided cascade remains slightly higher, indicating that explicit event localization can provide coarse boundaries but does not necessarily yield the best final edited video quality in Table~\ref{tab:main_results}. The causal variant is substantially worse at strict localization tolerances (Acc@0.1: 10.0 vs.\ 45.5), reflecting the difficulty of pinpointing the exact trigger boundary without future-block context. Importantly, this localization gap does not translate into a comparable quality gap in Tab.~\ref{tab:main_results}: our region-based VLM evaluator scores pre-event preservation and post-event editing on ground-truth segments and is therefore relatively insensitive to small shifts in the predicted edit-start frame. Coarse timing errors that still place the edit in roughly the right region are penalized only mildly, which is why the causal variant remains close to the bidirectional model in averaged editing quality (Tab.~\ref{tab:main_results}) while lagging in strict-tolerance localization (Tab.~\ref{tab:event_timing}). Precise boundary timing is thus a distinct axis from editing quality, and improving causal localization is an explicit direction for future work. 

\paragraph{Efficiency ablation.}

Table~\ref{tab:efficiency} compares inference cost on the same Ego4D event sample. EgoPlay has nearly the same latency as the VLM-guided cascade but uses less than half the GPU memory because it does not keep a separate VLM localizer resident. EgoPlay-Causal gives the best resource profile and is the only streamable variant, reducing memory from 45.96 GB to 41.87 GB relative to EgoPlay while maintaining comparable throughput. These results show that event-triggered editing can be integrated into the editor itself without introducing a separate detection model, and that causal inference is the preferred deployment path when streaming and memory are the primary constraints. Finally, Fig.~\ref{fig:self_forcing} shows qualitative results 
from a 4-step self-forcing distilled variant targeting real-time 
deployment; while successful event-triggered edits are possible 
at 16\,FPS, robustness remains limited and we leave this as an open direction.

\revision{To test streaming beyond the five-second training horizon, we also run the causal model on 35 clips extended to 11.6--14.9 seconds (\longhorizonref). With full KV context, temporal consistency remains flat from 0.743 within the training horizon to 0.752 beyond it, and the full-clip VLM score is 7.76 versus 7.39 for the five-second prefix. A constant-memory rolling cache reveals eviction-induced artifacts, motivating eviction-aware training for indefinite streams.}

\input{tables/efficiency}

%% file: tables/editversebench.tex
\begin{table}[t!]
\centering
\caption{Quantitative comparison on EgoEditBench and EditVerseBench. 
``PS'', ``TA'', and ``TC''  denote Pick Score, Text Alignment, and Temporal Consistency metrics. 
Reference-based editing tasks from EditVerseBench---propagation, inpainting, reference insertion,
 and edit with mask---were excluded. 
 ``$\dagger$'' indicates closed-source models evaluated using their publicly released samples; 
 ``$\ddagger$'' indicates models utilizing the first frame generated by EgoEdit. 
 EgoEdit-RT stands for the real-time streaming version of EgoEdit.}
 \label{tab:egoedit_comparison}
 \vspace{-1em}
\scriptsize
\setlength{\tabcolsep}{2.0pt}
\renewcommand{\arraystretch}{1.02}
\resizebox{\columnwidth}{!}{%
\begin{tabular}{llcccccccc}
\toprule
\multirow{2}{*}{\textbf{Method}} &
\multirow{2}{*}{\textbf{Family}} &
\multicolumn{4}{c}{\textbf{EgoEditBench}} &
\multicolumn{4}{c}{\textbf{EditVerseBench}} \\
\cmidrule(lr){3-6} \cmidrule(lr){7-10}
& & \textbf{VLM$\uparrow$} & \textbf{PS$\uparrow$} & \textbf{TA$\uparrow$} & \textbf{TC$\uparrow$}
  & \textbf{VLM$\uparrow$} & \textbf{PS$\uparrow$} & \textbf{TA$\uparrow$} & \textbf{TC$\uparrow$} \\
\midrule

TokenFlow & \multirow{2}{*}{Attention manipulation}
& 4.99 & \underline{18.91} & 15.89 & 95.04
& 5.87 & \textbf{19.90} & 23.68 & 98.21 \\
STDF
& & 4.59 & 18.69 & 15.64 & 93.96
& 6.64 & 19.54 & 24.33 & 96.96 \\

\midrule

Señorita-2M$^{\dagger}$
& \multirow{2}{*}{First-frame propagation}
& \underline{7.52} & 18.85 & 16.25 & \underline{95.86}
& 6.99 & 19.32 & 23.07 & 98.33 \\
AnyV2V$^{\ddagger}$
& & 6.72 & 18.65 & 15.35 & 92.37
& 6.46 & 19.47 & 23.32 & 95.91 \\

\midrule

InsV2V
& \multirow{4}{*}{Instruction-guided}
& 5.24 & 18.81 & 14.92 & 94.01
& 5.71 & 19.08 & 22.49 & 96.39 \\
Lucy Edit
& & 5.44 & 18.87 & 15.03 & 94.41
& 6.27 & 19.23 & 22.55 & \underline{98.62} \\
EditVerse$^{\dagger}$
& & -- & -- & -- & --
& \textbf{8.26} & \underline{19.69} & \textbf{25.29} & \textbf{98.68} \\
\textbf{EgoEdit}
& & \textbf{7.76} & \textbf{19.21} & \textbf{16.89} & \textbf{96.70}
& \underline{8.00} & 19.61 & \underline{24.40} & 98.54 \\
\rowcolor{blue!6}
\textbf{EgoPlay (ours)}
& & 7.60 & 18.93 & 15.93 & 94.92
& 7.64 & 19.67 & 24.26 & 98.44 \\    
\midrule

StreamDiffusion
& \multirow{3}{*}{Streaming models}
& \underline{4.32} & \underline{18.92} & \underline{14.15} & 86.83
& \underline{4.33} & 18.76 & \textbf{19.01} & 93.41 \\
StreamDiffusionV2
& & 2.55 & 18.63 & 12.75 & 94.31
& 2.78 & 18.45 & 17.32 & \underline{98.22} \\
\textbf{EgoEdit-RT}
& & \textbf{7.71} & \textbf{19.13} & \textbf{16.34} & \textbf{96.41}
& \textbf{8.18} & \textbf{19.59} & 17.61 & \textbf{98.55} \\
\rowcolor{blue!6}
\textbf{EgoPlay Causal (ours)}
& & 6.94 & 18.81 & 15.54 & 92.14
& 6.83 & 19.38 & 23.37 & 96.95 \\    
\bottomrule
\end{tabular}
}
\end{table}

%% file: tables/table_user_study.tex
\begin{table}[t]
\centering
\scriptsize
\caption{%
  \textbf{Human Preference Study.}
  Participants chose between EgoPlay and each baseline on four criteria.
  \textbf{W}~=~prefer ours, \textbf{T}~=~tie, \textbf{L}~=~prefer baseline
  (10 participants, 300 judgments per row and 1200 per baseline).
  Aggregated across criteria, EgoPlay's win rate exceeds 50\% against every
  baseline; the only per-criterion tie is edit quality vs.\ the
  near-equivalent EgoPlay-Causal variant.
}
\label{tab:userstudy}
\setlength{\tabcolsep}{5pt}
\begin{tabular}{l l ccc}
\toprule
\textbf{Baseline} & \textbf{Criterion} & \textbf{W\,(\%)} & \textbf{T\,(\%)} & \textbf{L\,(\%)} \\
\midrule
\multirow{5}{*}{VLM}
  & Instruction Following & 90 &  6 &  4 \\
  & Edit Timing           & 80 & 16 &  4 \\
  & Edit Quality          & 82 & 10 &  8 \\
  & Overall               & 90 &  4 &  6 \\
  \cmidrule(lr){2-5}
  & \textit{All criteria} & \textbf{86} &  9 &  6 \\
\midrule
\multirow{5}{*}{EgoEdit}
  & Instruction Following & 80 & 12 &  8 \\
  & Edit Timing           & 78 & 12 & 10 \\
  & Edit Quality          & 84 &  4 & 12 \\
  & Overall               & 84 &  8 &  8 \\
  \cmidrule(lr){2-5}
  & \textit{All criteria} & \textbf{82} &  9 & 10 \\
\midrule
\multirow{5}{*}{EgoPlay-Causal}
  & Instruction Following & 60 & 16 & 24 \\
  & Edit Timing           & 66 &  6 & 28 \\
  & Edit Quality          & 50 & 20 & 30 \\
  & Overall               & 64 &  6 & 30 \\
  \cmidrule(lr){2-5}
  & \textit{All criteria} & \textbf{60} & 12 & 28 \\
\bottomrule
\end{tabular}
\end{table}

%% file: tables/event_timing.tex
\begin{table}[t]
    \centering
    \footnotesize
    \setlength{\tabcolsep}{6pt}
    \renewcommand{\arraystretch}{1.12}
    \caption{Edit-start localization accuracy on positive Ego4D event-triggered samples. Acc@$r$ counts a prediction as correct if the visible edit starts within $rT$ frames of the ground-truth trigger end, where $T$ is the clip length. Higher is better.}
    \label{tab:event_timing}
    \vspace{-1em}
    \begin{tabular}{lccc}
    \toprule
    Method & Acc@0.1 $\uparrow$ & Acc@0.2 $\uparrow$ & Acc@0.3 $\uparrow$ \\
    \midrule
    VLM-guided & 40.0 & 64.0 & 72.0 \\
    \rowcolor{blue!6}
    EgoPlay (ours) & 45.5 & 63.6 & 68.2 \\
    \rowcolor{blue!6}
    EgoPlay-Causal (ours) & 10.0 & 15.0 & 60.0 \\
    \bottomrule
    \end{tabular}
    \end{table}

%% file: tables/efficiency.tex
\begin{table}[t]
\centering
\caption{Efficiency comparison on one positive Ego4D event sample. Each method is evaluated on an 81-frame $512{\times}288$ clip with batch size 1, averaged over multiple inference runs. VLM-guided memory sums the EgoEdit editor and Qwen3-VL localizer.}
\label{tab:efficiency}
\vspace{-1em}
\scriptsize
\setlength{\tabcolsep}{2.5pt}
\renewcommand{\arraystretch}{1.05}
\resizebox{\columnwidth}{!}{%
\begin{tabular}{lccccc}
\toprule
\textbf{Method} & \textbf{Time (s)$\downarrow$} & \textbf{FPS$\uparrow$} & \textbf{Throughput$\uparrow$} & \textbf{GPU Mem. (GB)$\downarrow$} & \textbf{Streamable} \\
\midrule
VLM-guided & 70.00 & 1.16 & 0.014 & 102.98 & No \\
EgoPlay & 67.82 & 1.19 & 0.015 & 45.96 & No \\
\rowcolor{blue!6}
\textbf{EgoPlay-Causal} & \textbf{66.59} & \textbf{1.22} & \textbf{0.015} & \textbf{41.87} & Yes \\
\bottomrule
\end{tabular}
}
\end{table}

%% file: sections/5_conclusion.tex
\section{Conclusion}

We introduced EgoPlay, an event-triggered video editing framework for egocentric streams. 
Unlike conventional video editors that apply an instruction to an entire clip or to a manually specified region, 
EgoPlay conditions editing on visual events: 
it must detect when the requested trigger occurs, 
preserve all pre-event content, and synthesize a coherent edited continuation after the trigger.
To support this setting, we built a large-scale event-triggered editing dataset with positive, negative, 
and multi-event prompts that train the model to decide both \emph{when} to edit and \emph{when not} to edit.
Experiments on Ego4D show that EgoPlay achieves the best averaged event-triggered editing performance among methods 
that do not receive ground-truth event boundaries, outperforming both EgoEdit and a VLM-guided detector--editor cascade. 
The results suggest that event-triggered generation is a promising direction for always-on egocentric and AR systems, 
where visual effects should respond to user behavior without manual temporal masks or post-hoc stitching.

Several challenges remain. The model must become more conservative on difficult negative prompts, 
more robust under long-horizon streaming, and more precise in timing edits for subtle or ambiguous events. 
Future work can extend EgoPlay with longer memory, stronger causal rollout training, interactive user correction, 
and real-time deployment in wearable or AR settings.

%% file: sections/acknowledgments.tex
\begin{acks}
\ifarxivbuild\else
Ashkan Mirzaei and Runjia Li contributed to the project through valuable discussions during its early stages.
\fi
Jinjie Mai conducted this work during his internship at Snap Inc.
This work was supported in part by the King Abdullah University of Science and Technology (KAUST) Center of Excellence for Generative AI under award 5940 and the SDAIA-KAUST Center of Excellence in Data Science and Artificial Intelligence.
\end{acks}

%% file: figures/comp_figure2.tex
\begin{figure*}[!t]
  \centering
   \includegraphics[width=\linewidth]{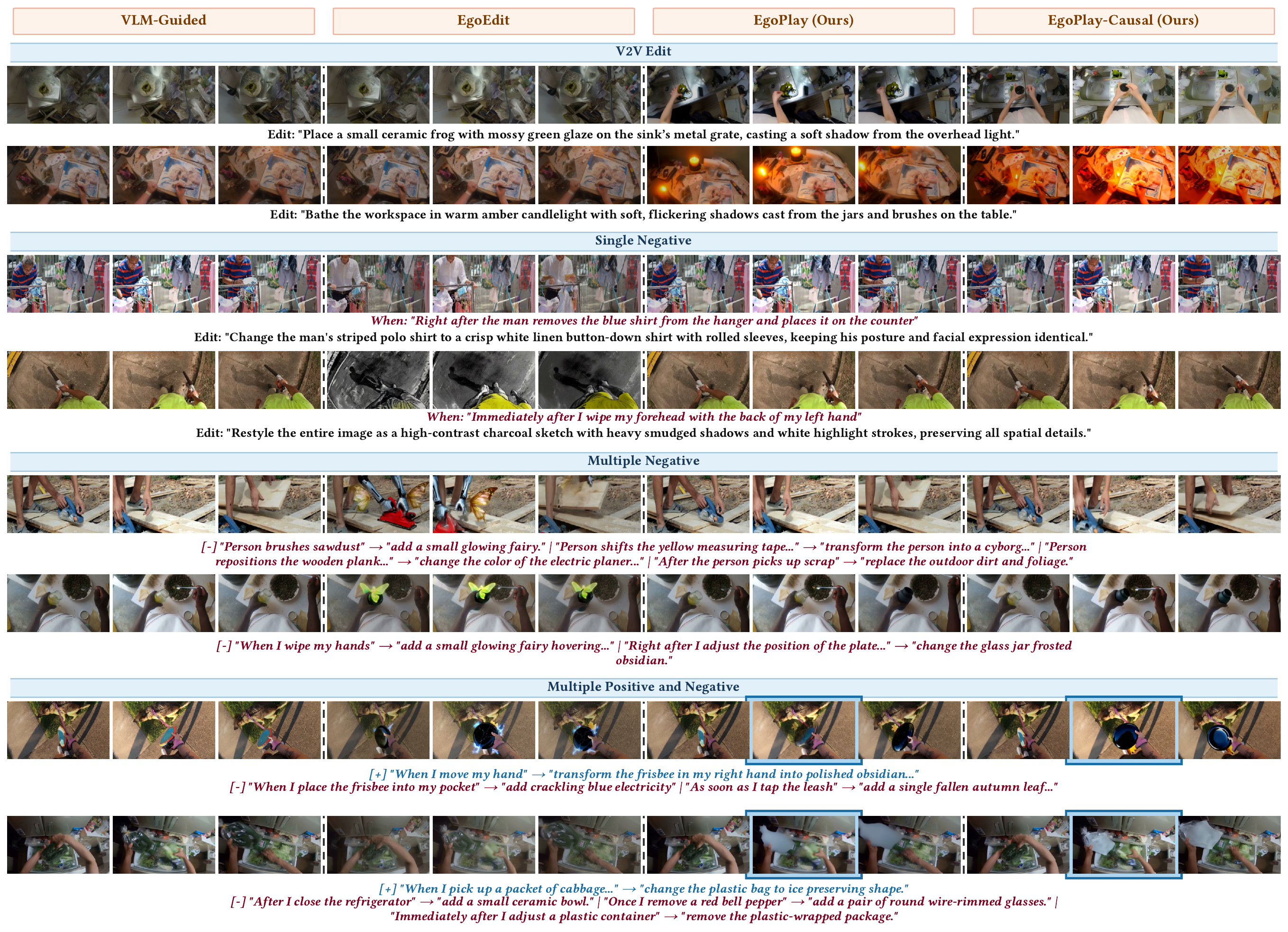}
   \caption{\textbf{Qualitative comparison across diverse evaluation modes.} We evaluate four methods---VLM-guided, EgoEdit, EgoPlay, and EgoPlay-Causal---on four increasingly challenging conditioning scenarios. (V2V Edit) Standard video editing with no temporal trigger. (Single Negative) The edit is conditioned on a single negative trigger, a moment when the edit should not apply. (Multiple Negative) Multiple negative triggers (maroon), each paired with a distinct distractor edit the model must ignore. (Multiple Positive and Negative) A mix of a true positive trigger (blue, \textbf{[+]}) that should activate the edit, alongside negative distractors (maroon, \textbf{[-]}) that should be suppressed. EgoPlay responds to positive triggers while remaining robust to negative distractors, whereas baselines often apply edits indiscriminately or exhibit artifacts.}
   \Description{Qualitative comparison of VLM-guided editing, EgoEdit, EgoPlay, and EgoPlay-Causal across V2V editing, single-negative, multiple-negative, and mixed positive-negative event-triggered scenarios.}
   \label{fig:comp_figure2}
\end{figure*}

%% file: figures/self_forcing.tex
\begin{figure*}[t]
    \centering
    \includegraphics[width=\linewidth]{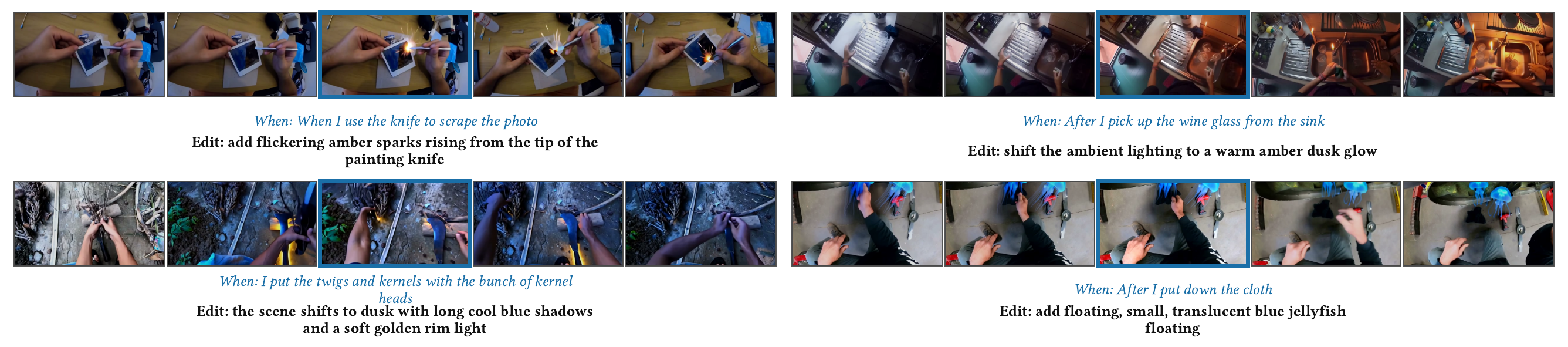}
    \caption{
    \textbf{EgoPlay-Causal-Realtime.}
    We further train a 4-step streaming model with self-forcing for real-time deployment.
    The model can produce successful event-triggered edits at 16 FPS (top row), but it is less robust than our 40-step causal model and can sometimes begin editing from the start of the clip (bottom row).
    }
    \label{fig:self_forcing}
\end{figure*}

%% file: sections/supp.tex
\section{Limitations}

EgoPlay introduces event-triggered video editing as a first step toward reactive egocentric and AR video generation, 
but several challenges remain. 
First, the causal model improves deployability by removing access to future frames, 
but precise event localization remains harder without future context, especially for subtle or ambiguous triggers. 
Improving causal memory and rollout training is therefore an important direction for more temporally precise streaming editing.

Our real-time streaming variant is also still an early prototype. 
The 4-step self-forcing distilled model can produce successful event-triggered edits at 16\,FPS, 
but its robustness is not yet comparable to the full model. 
In addition, EgoPlay can be less conservative than an explicit detector--copy pipeline on difficult negative prompts, 
and broader evaluation across domains and event ambiguity levels will be useful. 
These limitations do not change the main finding that joint event understanding and video editing
can outperform conventional editors and detector--editor cascades on the primary Ego4D benchmark without using ground-truth event boundaries;
they instead point to future work on temporal precision, negative-event restraint, and robust real-time deployment.

\subsection{Limitations of VLM-Based Evaluation}
Our benchmark scores edits with a VLM judge (Qwen3-VL-30B~\cite{bai2025qwen3}),
which enables scalable, reference-free evaluation but is not yet a fully reliable grader:
current VLMs occasionally assign high scores to clearly incorrect outputs.
\cref{fig:vlm_eval_limitation} shows a multiple-negative sample in which the
fabricated trigger never occurs---so the correct behavior is to leave the video
unedited---yet both EgoPlay and EgoPlay-Causal hallucinate the conditioned edit
while the evaluator still awards a perfect $9/9$. Such false positives can
inflate negative-mode scores and partly explain the residual gap on negative
settings. We expect stronger future VLMs to largely close this gap; in the
meantime, our benchmark contributes a standardized, reproducible event-triggered
editing protocol and dataset that make these failure modes measurable and that
can be re-scored as better judges become available.
\input{figures/vlm_eval_limitation}

\section{Supplementary Results}

\subsection{Auxiliary Common-Video Split}
The main paper reports event-conditioned editing results on Ego4D, which is our primary benchmark because it directly matches the egocentric, first-person setting targeted by EgoPlay. In this supplement, we additionally report results on an auxiliary common-video split. This split contains general-domain videos and is used primarily as a small diversity source during training, rather than as the main benchmark for the paper.

The auxiliary split is useful as a diagnostic because it contains more varied scene layouts, camera motions, and object categories than Ego4D. However, it is less aligned with the intended egocentric use case and contains fewer evaluation samples. We therefore use it to test robustness and failure modes, while keeping the main-paper claims focused on Ego4D.

\input{tables/common_results_supp}

\paragraph{Quantitative and Qualitative results.}
Table~\ref{tab:common_results_supp} reports the auxiliary common-video results under the same four event-conditioned modes used in the main Ego4D evaluation. On the single-positive setting, where a real trigger is present and the model must apply the requested edit after that event, EgoPlay and EgoPlay-Causal substantially outperform the direct EgoEdit baseline. EgoEdit obtains a total score of 5.52, while EgoPlay improves to 6.88 and EgoPlay-Causal reaches 7.07. This indicates that event-conditioned training transfers beyond Ego4D-style first-person videos and helps the model couple trigger recognition with post-event editing.

The multiple-positive-and-negative setting is more challenging because the model must identify the true trigger while ignoring distractor events and edits. On this setting, EgoPlay and EgoPlay-Causal remain competitive with the VLM-guided cascade: EgoPlay obtains 5.05, EgoPlay-Causal obtains 5.08, and VLM-guided obtains 5.34. These results suggest that compositional event-conditioned prompting improves robustness to distractors, although explicit detector--editor cascades can still be advantageous when the detector correctly rejects irrelevant conditions.

The negative modes reveal a complementary behavior. VLM-guided editing obtains the strongest scores on single-negative and multiple-negative samples because it can explicitly reject the fabricated trigger and copy the input video. EgoPlay preserves many no-edit cases, but its end-to-end generative formulation is less conservative than a detector--copy pipeline. This explains why VLM-guided has the highest average score on the auxiliary common-video split, even though EgoPlay variants are stronger on the positive trigger setting.

Fig.~\ref{fig:common_comp_figure} shows qualitative comparisons on 
the auxiliary common-video split examples in 
the single-positive setting.
\input{figures/comp_figure_common}

\paragraph{Takeaway.}
These supplementary results do not change the main conclusion of the paper. On the primary Ego4D benchmark, EgoPlay achieves the best averaged event-conditioned score among methods that do not receive ground-truth event boundaries. The auxiliary common-video split instead provides additional evidence about transfer beyond egocentric videos and highlights a remaining challenge: end-to-end generative editors should become more conservative when prompts describe plausible but absent events.

\ifcamerareview\color{blue}\fi
\section{Data-Generation Reproducibility}

\subsection{Ego4D Filtering and Event Selection}
We apply all quality filters before event and prompt generation. The default thresholds used for the released training split are: aesthetic score (>3.5), DOVER score (>0.3), FFmpeg blur mean (<6.4), VMAF motion score (<20), brightness in ([60,130]), saturation (>5), no detected black border, jitter score (<20), and mean LPIPS trajectory (<0.2). Missing or failed measurements are rejected. We initially retain at most one sampled clip per Ego4D video and later cap the enriched set at four clips per source video to limit repeated scenes.

A text model first accepts only a single, visible physical action with a crisp completion point. It rejects camera motion, ambient events, static or continuous states, vague descriptions, and multi-step actions. Qwen3-VL-235B then receives the start, midpoint, and end frames, rewrites the narration in first person, and assigns an alignment score from 1 to 3. We retain only score-3 events. Each training window has sufficient pre-event, event, and post-event frames; trigger positions are rebalanced after window sampling rather than always placing the event at the same offset.

\subsection{Prompt Construction}
For each event, Qwen3-VL-235B observes one post-event frame and first selects feasible types from a catalog of 21 edits. Four feasible types are sampled. The generation prompt requires one first-person sentence per edit, beginning with a trigger clause, grounded only in visible objects, and returned as ordered JSON:
{\scriptsize
\begin{quote}
\texttt{Event: [first-person event]}\\
\texttt{Requested edit types: [ordered list]}\\
\texttt{Return \{"edit\_prompts": [}\\
\texttt{  \{"edit\_type": ..., "prompt": ...\}, ...]\}.}
\end{quote}
}
The system instruction requires the edit to begin immediately after event completion, forbids placeholders and additional edits, and requires the output list to match the requested types and order. We then deterministically sample one of 24 condition forms, including ``After [event], [edit],'' ``When [event] happens, [edit],'' and ``Only after [event] should [edit].'' A text-only rewrite removes the trigger clause and resolves dangling pronouns to form the event-free synthesis instruction.

For each positive, a second two-frame prompt fabricates two plausible but absent single-action triggers. The first preserves the positive edit and changes only the event; the second uses a different event and edit type. The JSON schema is:
{\scriptsize
\begin{quote}
\texttt{\{"negative\_event\_edit\_prompt":}\\
\texttt{  [same edit, absent event],}\\
\texttt{ "negative\_event\_edit\_prompt\_2":}\\
\texttt{  [new edit, different absent event]\}.}
\end{quote}
}
Multi-event prompts concatenate one positive or negative anchor with one to three negative distractors and randomize the anchor position. The benchmark and evaluator prompts are reproduced below.

\subsection{Edited-Target Synthesis}
The source segment contains the pre-event, event, and post-event footage. We run the pretrained EgoEdit V2V model on the post-event segment with the event-free instruction. At the boundary, the first eight source and edited frames are encoded by the VAE from \texttt{Qwen/Qwen-Image-2512}; corresponding latents are spherically interpolated and decoded to obtain eight transition frames. The positive target concatenates the untouched prefix and event, these transition frames, and the remaining edited suffix. A negative target is exactly the source. Generated pairs are filtered again by Qwen3-VL-235B for editing quality, visual quality, and background consistency before training.

\section{Additional Experimental Analysis}

\subsection{Additional Open-Source Baselines}
\input{tables/additional_baselines}
Ditto uses Wan2.1-VACE-14B with the released \texttt{ditto\_local} LoRA; VIVA uses its HunyuanVideo \texttt{ema-checkpoint-12000}. Both receive the event-free instruction and process the entire 81-frame clip, because neither exposes an event trigger. We use 50 denoising steps and seed 42, generate at each method's native resolution, resize frame-preservingly to $512\times288$, and score with the same Qwen3-VL-30B-A3B evaluator. These experiments use the submitted 40-pair set: 40 samples per mode and 200 outputs across five modes. Approximately 124 of the 400 baseline outputs ended slightly before frame 81; the evaluator padded the empty tail with its standard black placeholder rather than dropping samples, affecting both methods symmetrically.

\subsection{Commercial System Comparison}
\input{tables/runway_results}
We manually evaluate five positive clips using the same source, event instruction, length, frame rate, and annotated boundary. Runway Aleph2 agent mode orchestrates event localization, appearance editing, video generation, and stitching; Runway V2V is a single event-unaware edit. Aleph2 obtains the highest VLM score, but its mean JUMP is $0.398$, $1.73\times$ the source's natural-motion value of $0.230$, and its temporal consistency indicates stronger stabilization than the source. Because this is a positive-only $n=5$ spot check with one VLM scoring run, we use it as qualitative commercial context rather than a population-level ranking.

\subsection{Long-Horizon Analysis}
\input{tables/long_horizon}
We extend 35 positive benchmark clips with continuous footage from the same Ego4D recording; five clips lack sufficient continuation and are excluded. The event occurs in the first five seconds and its edit must persist for the remaining 7--10 seconds. Full context retains all past blocks and reaches 66.6 GB at roughly 15 seconds. Rolling-11 evicts older blocks, remains at 42.5 GB, and is 13\% faster per block, but eviction was never encountered during five-second training and causes high-frequency and color artifacts. The full-context result therefore demonstrates finite long-horizon stability, while eviction-aware training remains necessary for robust indefinite constant-memory streaming. VLM totals are from one run and exhibit approximately \(\pm0.2\) variation; source tracking also conflates a persistent edit with tracking error.

\subsection{User-Study Interface}
\input{figures/user_study_interface}
All ten participants received the same 30 trials. Output order was randomized and method names were hidden. The reference was produced with the human event boundary: pre-event and event frames were copied from the source and only the post-event suffix was edited. Participants compared each anonymized output against this target for instruction following, edit timing, edit quality, and overall preference. Figure~\ref{fig:user_study_interface} illustrates the interface populated with synchronized frames from the second single-positive benchmark example.

\section{Implementation Details}

All event-triggered editing models operate on 81-frame clips at \(512 \times 288\) resolution.
Before diffusion, videos are encoded into a latent representation with \(8\times\) spatial compression and \(4\times\) temporal compression, so an 81-frame clip becomes 21 latent frames of spatial size \(64 \times 36\).
The source and target latents each use 16 channels and are injected through channel-wise conditioning as described in the main paper.

For the causal EgoPlay variant, we split the 21-frame latent sequence into \(N=7\) temporal blocks.
Each block contains 3 latent frames.
Visual self-attention is bidirectional inside each block but causal across blocks: block \(B_i\) can attend to blocks \(B_{\leq i}\), and cannot attend to future blocks.
This block structure is used both during causal training and streaming-style inference.

\paragraph{Bidirectional training.}
We initialize from the pretrained 40-step EgoEdit V2V model and fine-tune for 30,000 steps in bfloat16. We use fused AdamW with learning rate $3\times10^{-5}$, weight decay 0.01, $\beta=(0.9,0.99)$, gradient clipping at 1.0, a cosine schedule with 1,000 warmup steps and minimum learning rate $10^{-5}$, and EMA decay 0.9999. Training samples contain 81 frames at $512\times288$; the active per-stream video batch size is 2.

\paragraph{Causal training.}
The causal model initializes from bidirectional step 20,000 and is fine-tuned for 30,000 steps with three latent frames per block, 21 latent frames total, and the same bfloat16 precision, optimizer betas, weight decay, clipping, and EMA decay. Its learning rate is $10^{-5}$, with 500 warmup steps, a cosine schedule, and minimum learning rate $10^{-6}$. Both full models use 40 denoising steps at inference.

\paragraph{Real-time distillation.}
The self-forcing student generates in four denoising steps at times $(1.0,0.9,0.75,0.5)$, uses three latent frames per block, and trains on 21 latent frames. It reaches 16 FPS in our prototype. As shown in the qualitative failure case, it can trigger correctly but may also edit from the first frame; we therefore report it as a real-time pathway rather than claim production robustness.

\paragraph{Release.}
We will release the evaluation benchmark videos, annotations, prompts, and scoring protocol through the project page. This commitment covers the benchmark; it does not claim release of proprietary training footage, model weights, or the commercial EgoEdit implementation.

\section{Evaluation Details}

\subsection{VLM-Guided Detector--Editor Baseline}
The VLM-guided baseline factorizes event-triggered editing into explicit event localization followed by conventional video editing.
We use Qwen3-VL-8B~\cite{bai2025qwen3} for this detector stage.
Given a source clip and an event-conditioned instruction,
the VLM performs a single video-language pass that decides whether any referenced trigger event occurs, 
estimates its start and end timestamps, and rewrites the request into an event-free edit prompt.
For positive and multi-event prompts, a detected trigger defines the point after which editing may begin; 
for negative prompts, the correct VLM behavior is to reject the fabricated trigger.

When the VLM detects an event, we extract an 81-frame source clip immediately following the predicted endpoint.
EgoEdit processes this clip using the resulting event-free prompt.
The final baseline output is then constructed by preserving the original source frames 
up to the predicted event endpoint and hard-cutting to the edited post-event result afterward.
If the VLM does not detect an event, the baseline rejects the condition and copies the source video unchanged.
Thus, the baseline tests whether an explicit VLM localizer and prompt rewriter 
can solve the task without training a single model to jointly perform event recognition and editing.

\paragraph{VLM localizer prompt.}
The detector stage uses the following prompt template with the source video and the original event-conditioned instruction:

{\small
\begin{quote}
You are given a video and an event-conditioned editing instruction.

\textbf{Instruction:} \emph{[event-conditioned instruction]}

Watch the entire video carefully. The video is \emph{[duration]} seconds long (\emph{[number of frames]} frames at \emph{[fps]} FPS).

\textbf{Task:}
1. Decide if any trigger event referenced by the instruction occurs in the video.
2. If yes, return the start and end timestamps of one matched event.
3. Return a concise no-event-condition edit prompt that should be applied after the event.

\textbf{Rules:}
If no event is detected, set the start and end timestamps to null.
If no event is detected, the rewritten edit prompt can be null.
Return only valid JSON:\\
\{\ ``event\_detected'': true/false,\\
\hphantom{\{}``start\_seconds'': float or null,\\
\hphantom{\{}``end\_seconds'': float or null,\\
\hphantom{\{}``event\_description'': string,\\
\hphantom{\{}``event\_trigger\_phrase'': string,\\
\hphantom{\{}``vlm\_model\_edit\_prompt'': string or null \}.
\end{quote}
}

\subsection{VLM-Based Evaluation Protocol}
We evaluate generated videos with Qwen3-VL-30B-A3B~\cite{bai2025qwen3}.
The evaluator compares the source video with the predicted video and assigns three scores from 0 to 3: 
editing quality, visual quality, and background consistency.
The total score is the sum of these three dimensions.
The evaluator uses the ground-truth event boundary only to decide which temporal regions should be edited; 
it does not use the method's predicted boundary for scoring.

For each sample, we split the source and prediction into semantically meaningful temporal regions.
For a standard video-to-video edit, all regions should contain the requested edit.
For a positive event-conditioned sample, the pre-event and event regions should remain unchanged, 
while the post-event region should contain the requested edit.
For a negative event-conditioned sample, the trigger is absent, so all regions should remain unchanged.
In regions that should be edited, the evaluator is shown the canonical event-free edit instruction and judges whether the output fulfills it.
In regions that should not be edited, the edit instruction is omitted, 
and the evaluator instead judges whether the prediction correctly preserves the source.
This prevents the evaluator from rewarding prompt agreement in regions where no edit should appear.

In the local video-based evaluator, each temporal region is represented by a short source--prediction clip pair.
If a region is too short for robust video-model sampling, it is padded to a minimum duration before evaluation.
Each region is scored independently, and the final sample score is a weighted average over regions.
For positive event-conditioned samples, the pre-event and event preservation regions each receive half weight, 
and the post-event editing region receives full weight, giving equal total importance to 
``do not edit before the trigger'' and ``edit after the trigger.''
For standard editing and negative samples, the temporal regions are weighted uniformly.

\paragraph{Evaluator prompt.}
For each source--prediction clip pair, the VLM receives the two clips and the following instruction template.
The first block is included only when the region should contain the edit; it is omitted for regions that should remain unchanged.

{\small
\begin{quote}
\textbf{Editing Prompt:} \emph{[event-free edit instruction, only for regions that should be edited]}

You are a meticulous video editing quality evaluator.
You are given two short video clips: the ``Before'' clip is the original source, and the ``After'' clip is the model output.
If an editing prompt is provided, evaluate how well the editing prompt has been executed in the ``After'' clip relative to the ``Before'' clip.
If no editing prompt is provided, evaluate whether the ``After'' clip was correctly left unchanged relative to the ``Before'' clip.

Evaluate across three criteria, each from 0 (worst) to 3 (best):

\textbf{1. Editing Quality.}
If the region should be edited, score whether the edit accurately and completely follows the editing prompt.
If the region should remain unchanged, score whether the model correctly left the clip unchanged.

\textbf{2. Visual Quality.}
Score whether the output is realistic, seamless, and free of artifacts such as blurriness, distortion, or unnatural textures.

\textbf{3. Background Consistency.}
Score whether areas that should not be edited remain stable and unchanged between the source and output.

Respond with only a JSON object containing the three numerical scores.
\end{quote}
}

\subsection{Edit-Timing Annotation Interface}
To analyze \emph{when} each method begins editing, we use the frame-level
annotation interface in \cref{fig:transition_and_annotation}: each predicted clip
is expanded into individual frames, and annotators mark the first frame with a
visible edit, with the ground-truth trigger-end overlaid for reference.

We further observe that EgoPlay does not switch to the edited appearance
abruptly, but learns an \emph{edit transition effect} in which the edit emerges
gradually over a few frames after the trigger. This follows directly from the
transition frames $h$ injected into $T^{+}$ during data synthesis
(\datasynthref), which EgoPlay reproduces at inference
without test-time transition supervision.

\input{figures/transition_and_annotation}

%% file: figures/vlm_eval_limitation.tex
\begin{figure*}[t]
  \centering
  \includegraphics[width=\linewidth]{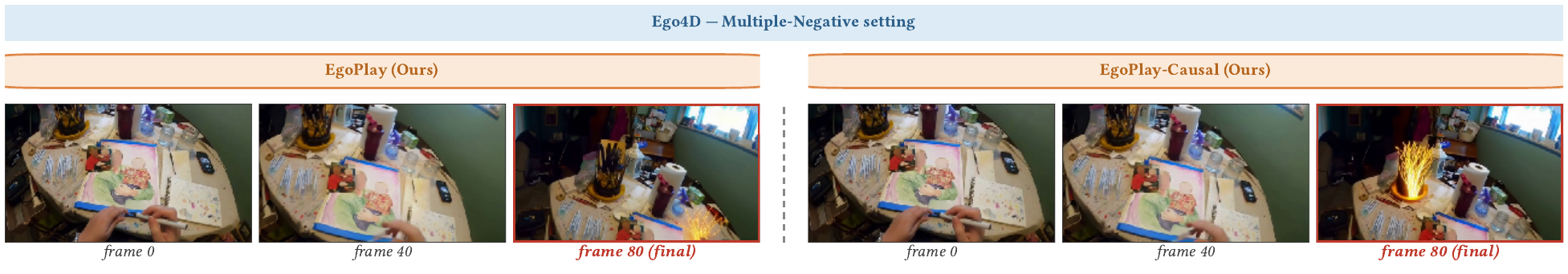}
  \caption{\textbf{Limitation of VLM-based evaluation.} A multiple-negative
  Ego4D sample. The conditioned edit (\textit{``transform the paint brush into a
  cluster of thin, glowing amber vines''}) is gated on a fabricated trigger
  (\textit{``once I pick up the photo of the child and place it on the
  notebook''}) that never occurs, so the correct behavior is to leave the video
  unedited. Both EgoPlay and EgoPlay-Causal instead hallucinate the edit by the
  final frame (red border), yet the Qwen3-VL-30B
  evaluator~\cite{bai2025qwen3} awards both a perfect $9/9$, showing that current
  VLM judges can reward edits that should not occur.}
  \Description{A multiple-negative example where both EgoPlay variants apply an
  edit that should not appear, while the VLM evaluator still gives a perfect
  score.}
  \label{fig:vlm_eval_limitation}
\end{figure*}

%% file: tables/common_results_supp.tex
\begin{table}[t]
\centering
\scriptsize
\setlength{\tabcolsep}{4pt}
\caption{Supplementary event-triggered editing results on the auxiliary common-video split (100 examples). We report editing quality, visual quality, background consistency, and total score. Higher is better.}
\label{tab:common_results_supp}
\begin{tabular}{llcccc}
\toprule
Mode & Method & \shortstack{Editing\\quality} & \shortstack{Visual\\quality} & \shortstack{Background\\consistency} & Total \\
\midrule
\shortstack[l]{Single\\positive}
& EgoEdit & 1.84 & 1.86 & 1.82 & 5.52 \\
& VLM-guided & 1.84 & 1.84 & 1.95 & 5.63 \\
& EgoPlay & 2.27 & 2.32 & 2.29 & 6.88 \\
& EgoPlay-Causal & \textbf{2.30} & \textbf{2.43} & \textbf{2.34} & \textbf{7.07} \\
\midrule
\shortstack[l]{Multiple positive\\and negative}
& EgoEdit & 1.05 & 1.15 & 1.16 & 3.35 \\
& VLM-guided & \textbf{1.66} & 1.75 & \textbf{1.94} & \textbf{5.34} \\
& EgoPlay & \textbf{1.66} & 1.73 & 1.66 & 5.05 \\
& EgoPlay-Causal & \textbf{1.66} & \textbf{1.77} & 1.66 & 5.08 \\
\midrule
\shortstack[l]{Single\\negative}
& EgoEdit & 1.73 & 1.82 & 1.76 & 5.31 \\
& VLM-guided & \textbf{2.87} & \textbf{2.87} & \textbf{2.87} & \textbf{8.60} \\
& EgoPlay & 2.60 & 2.69 & 2.64 & 7.93 \\
& EgoPlay-Causal & 2.20 & 2.31 & 2.20 & 6.71 \\
\midrule
\shortstack[l]{Multiple\\negative}
& EgoEdit & 0.94 & 1.18 & 1.07 & 3.19 \\
& VLM-guided & \textbf{3.00} & \textbf{3.00} & \textbf{3.00} & \textbf{9.00} \\
& EgoPlay & 2.31 & 2.56 & 2.31 & 7.19 \\
& EgoPlay-Causal & 2.06 & 2.17 & 2.06 & 6.29 \\
\midrule
\shortstack[l]{Average\\over modes}
& EgoEdit & 1.39 & 1.50 & 1.45 & 4.34 \\
& VLM-guided & \textbf{2.34} & \textbf{2.36} & \textbf{2.44} & \textbf{7.14} \\
& EgoPlay & 2.21 & 2.33 & 2.22 & 6.76 \\
& EgoPlay-Causal & 2.06 & 2.17 & 2.06 & 6.29 \\
\bottomrule
\end{tabular}
\end{table}

%% file: figures/comp_figure_common.tex
\begin{figure*}[t!]
  \centering
   \includegraphics[width=\linewidth]{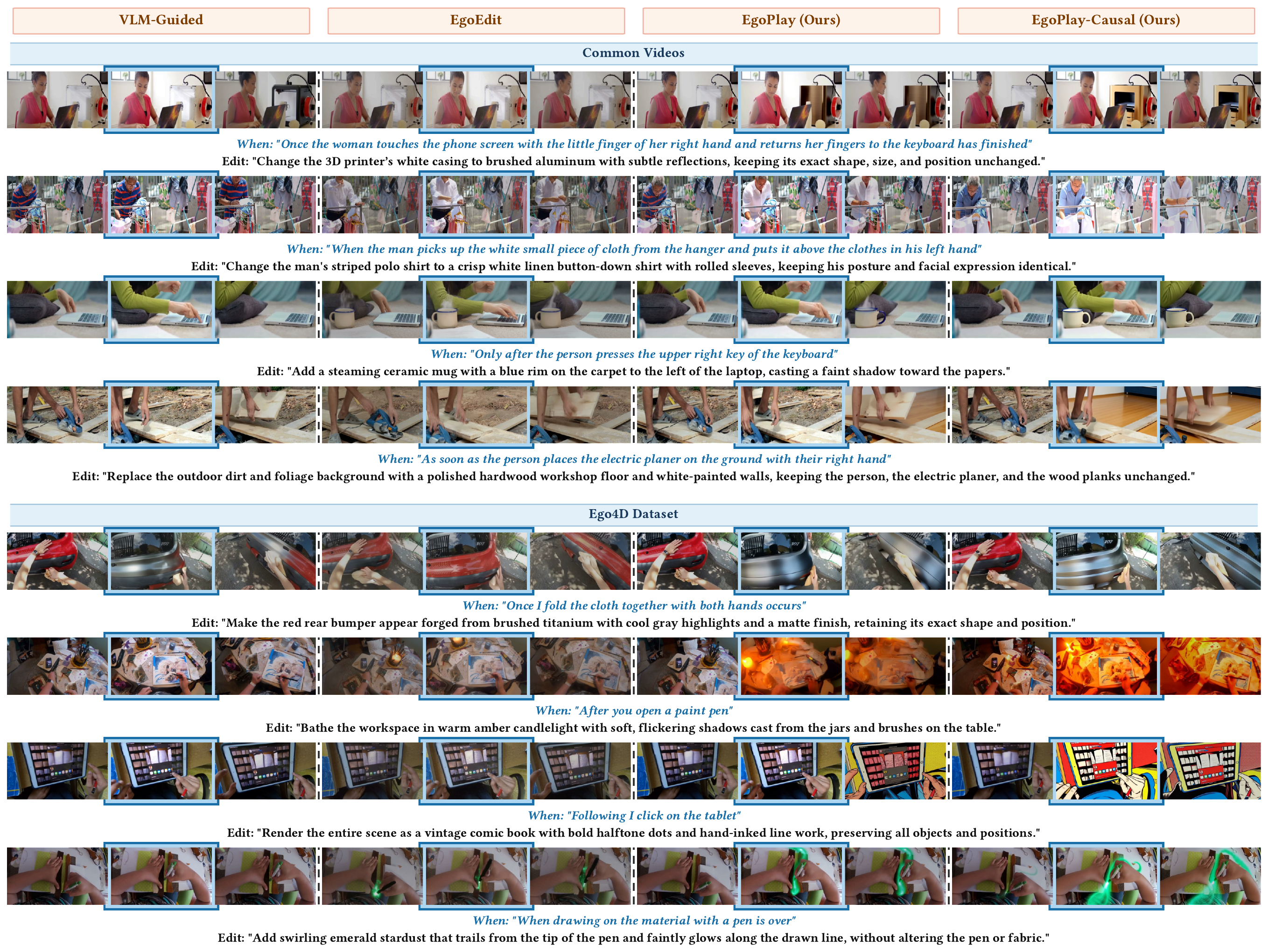}
      \vspace{-2em}
   \caption{\textbf{Qualitative comparison on the auxiliary common-video split (top) and Ego4D (bottom).} We compare EgoPlay and EgoPlay-Causal against VLM-guided editing and EgoEdit in the single-positive setting. For each example, three frames are shown: before, during (highlighted), and after the edit. Each edit is conditioned on a temporal trigger (When, italic) specifying the moment the edit should take effect, along with a spatial instruction (Edit, bold). EgoPlay better respects the trigger timing while maintaining edit fidelity and temporal coherence on this auxiliary split.}
   \Description{Qualitative comparison on the auxiliary common-video split for event-triggered video editing across EgoPlay, EgoPlay-Causal, VLM-guided editing, and EgoEdit.}
   \label{fig:common_comp_figure}
\end{figure*}

%% file: tables/additional_baselines.tex
\begin{table}[H]
\ifcamerareview\color{blue}\fi
\centering
\scriptsize
\caption{Additional open-source baselines on the submitted 40-pair Ego4D benchmark. Values are total VLM scores (0--9); the average covers the four event-triggered modes. All methods use their native prompt format and the same evaluator.}
\label{tab:additional_baselines}
\setlength{\tabcolsep}{3.2pt}
\begin{tabular}{lccccc}
\toprule
Method & Single pos. & Multi pos./neg. & Single neg. & Multi neg. & Avg. \\
\midrule
Ditto~\cite{bai2025ditto} & 4.93 & 4.88 & 4.43 & 4.65 & 4.72 \\
VIVA~\cite{cong2026viva} & 5.51 & 5.40 & 6.08 & 6.08 & 5.77 \\
EgoPlay & \textbf{7.56} & \textbf{7.14} & \textbf{8.21} & \textbf{7.95} & \textbf{7.72} \\
EgoPlay-Causal & 7.25 & 6.79 & 7.78 & 6.80 & 7.15 \\
\bottomrule
\end{tabular}
\end{table}

%% file: tables/runway_results.tex
\begin{table}[H]
\ifcamerareview\color{blue}\fi
\centering
\scriptsize
\caption{Commercial-system spot check on five positive benchmark clips. JUMP measures the worst adjacent-frame discontinuity relative to typical motion.}
\label{tab:runway_results}
\setlength{\tabcolsep}{3.4pt}
\begin{tabular}{lccc}
\toprule
Method & VLM total $\uparrow$ & Temp. consistency & JUMP $\downarrow$ \\
\midrule
Runway Aleph2 agent & \textbf{8.10} & 0.706 & 0.398 \\
EgoPlay-Causal & 7.65 & 0.643 & 0.264 \\
Runway V2V & 6.75 & 0.623 & 0.261 \\
Source reference & -- & 0.556 & 0.230 \\
\bottomrule
\end{tabular}
\end{table}

%% file: tables/long_horizon.tex
\begin{table}[H]
\ifcamerareview\color{blue}\fi
\centering
\scriptsize
\caption{Long-horizon causal generation on 35 positive clips extended to 11.6--14.9 seconds. ``Within'' is the first 5 seconds; ``beyond'' is the extrapolated suffix for pixel metrics, while ``full'' is the complete clip for VLM scoring.}
\label{tab:long_horizon}
\setlength{\tabcolsep}{3.5pt}
\begin{tabular}{llcccc}
\toprule
Cache & Region & Source track $\uparrow$ & Temp. cons. $\uparrow$ & Sharp ratio $\to1$ & VLM total $\uparrow$ \\
\midrule
Full context & within & 0.818 & 0.743 & 1.00 & 7.39 \\
Full context & beyond/full & 0.715 & \textbf{0.752} & 1.11 & \textbf{7.76} \\
Rolling-11 & within & 0.829 & 0.754 & 1.00 & 7.18 \\
Rolling-11 & beyond/full & 0.680 & 0.712 & 1.94 & 7.15 \\
\bottomrule
\end{tabular}
\end{table}

%% file: figures/user_study_interface.tex
\begin{figure*}[t]
\centering
\begingroup
\color{black}
\setlength{\fboxsep}{5pt}
\setlength{\fboxrule}{0.6pt}
\fcolorbox{egoplayline}{egoplaypanel}{%
\begin{minipage}{0.94\textwidth}
\small
\colorbox{egoplayorange!18}{%
\parbox{\dimexpr\linewidth-2\fboxsep\relax}{%
\textcolor{egoplayorangeDark}{\textbf{EVENT-TRIGGERED EDIT}}\quad
If and when I place a decoration on the window frame, change the decoration's material to iridescent mother-of-pearl with fine concentric grooves catching the sunlight exactly as before.}}
\par\medskip

\centering
\textcolor{egoplayorangeDark}{\textbf{REFERENCE TARGET}}\par\vspace{2pt}
\fcolorbox{egoplayorange}{white}{%
\includegraphics[width=0.29\linewidth]{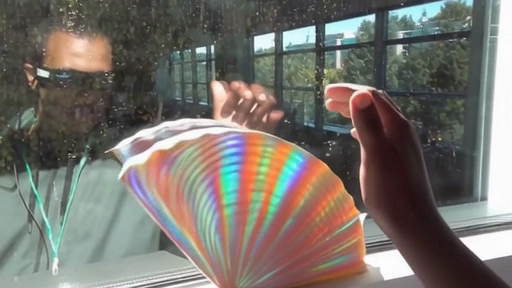}}
\par\medskip

\setlength{\tabcolsep}{2.5pt}
\begin{tabular}{@{}cccc@{}}
\colorbox{egoplayblue!18}{\makebox[0.218\linewidth]{\textcolor{egoplayblueDark}{\textbf{OUTPUT A}}}} &
\colorbox{egoplayblue!18}{\makebox[0.218\linewidth]{\textcolor{egoplayblueDark}{\textbf{OUTPUT B}}}} &
\colorbox{egoplayblue!18}{\makebox[0.218\linewidth]{\textcolor{egoplayblueDark}{\textbf{OUTPUT C}}}} &
\colorbox{egoplayblue!18}{\makebox[0.218\linewidth]{\textcolor{egoplayblueDark}{\textbf{OUTPUT D}}}} \\
\fcolorbox{egoplayblue}{white}{\includegraphics[width=0.205\linewidth]{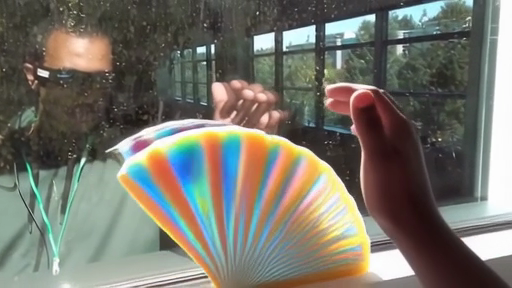}} &
\fcolorbox{egoplayblue}{white}{\includegraphics[width=0.205\linewidth]{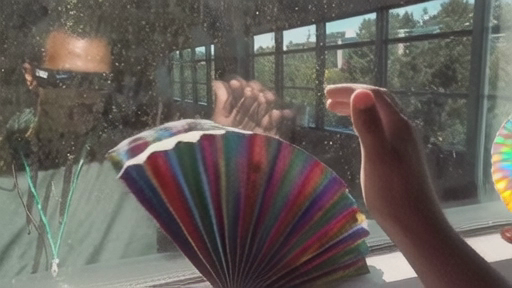}} &
\fcolorbox{egoplayblue}{white}{\includegraphics[width=0.205\linewidth]{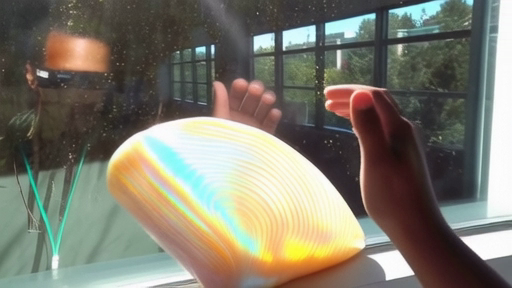}} &
\fcolorbox{egoplayblue}{white}{\includegraphics[width=0.205\linewidth]{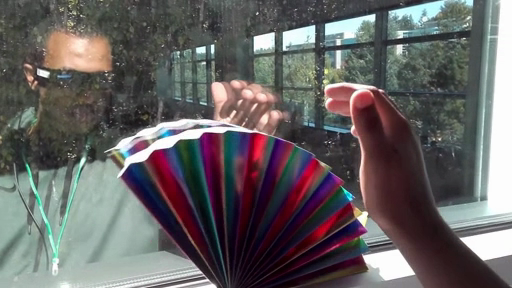}}
\end{tabular}
\par\medskip

\setlength{\fboxsep}{4pt}
\fcolorbox{egoplayline}{white}{%
\begin{minipage}{0.64\linewidth}
\centering
\textcolor{egoplayblueDark}{\textbf{SELECT THE BEST OUTPUT FOR EACH CRITERION}}
\par\vspace{2pt}
\renewcommand{\arraystretch}{1.08}
\setlength{\tabcolsep}{8pt}
\begin{tabular}{@{}lccccc@{}}
\rowcolor{egoplayblue!18}
 & \textbf{A} & \textbf{B} & \textbf{C} & \textbf{D} & \textbf{Tie} \\
Instruction following & \(\circ\) & \(\circ\) & \(\circ\) & \(\circ\) & \(\circ\) \\
\rowcolor{egoplaypanel}
Edit timing & \(\circ\) & \(\circ\) & \(\circ\) & \(\circ\) & \(\circ\) \\
Edit quality & \(\circ\) & \(\circ\) & \(\circ\) & \(\circ\) & \(\circ\) \\
\rowcolor{egoplaypanel}
Overall preference & \(\circ\) & \(\circ\) & \(\circ\) & \(\circ\) & \(\circ\)
\end{tabular}
\end{minipage}}
\end{minipage}}
\endgroup
\ifcamerareview\color{blue}\fi
\caption{\textbf{User-study interface.} Representative frames from the second single-positive benchmark example populate the interface. The study displayed synchronized reference and output videos rather than still images. Method identities were hidden, and the four outputs were randomly ordered for each trial.}
\label{fig:user_study_interface}
\end{figure*}

%% file: figures/transition_and_annotation.tex
\begin{figure*}[t!]
  \centering
  \includegraphics[width=\linewidth]{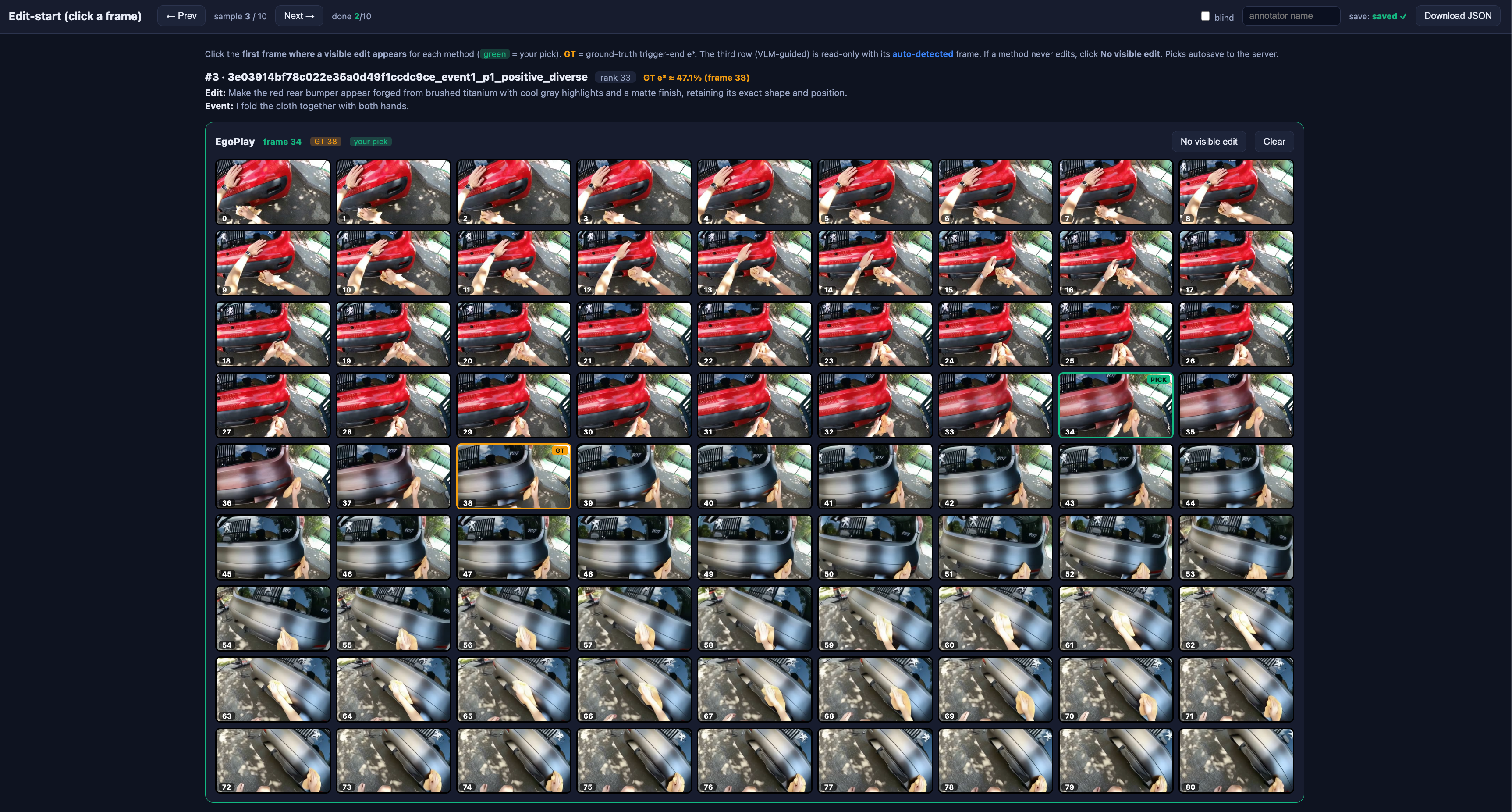}
  \caption{\textbf{Edit-timing annotation interface and learned edit transition.}
  Annotators mark the first edited frame in each predicted video, with the
  ground-truth trigger-end as reference. EgoPlay learns a smooth edit transition
  after the trigger rather than a hard cut, consistent with the transition frames
  used during data synthesis (see \datasynthref).}
  \Description{Screenshot of the frame-level edit-timing annotation interface
  alongside an example of EgoPlay's learned no-edit-to-edit transition after the
  trigger event.}
  \label{fig:transition_and_annotation}
\end{figure*}